\documentclass[runningheads]{llncs}
\usepackage[T1]{fontenc}
\usepackage{graphicx}
\usepackage{subcaption}  %
\usepackage{url}  %

\usepackage{wrapfig}

\usepackage{algorithm}
\usepackage{algorithmic}
\usepackage{booktabs} %
\usepackage{multirow} %

\usepackage{amsmath}
\usepackage{enumitem}
\usepackage{stmaryrd}
\usepackage{amssymb}
\usepackage{mathtools}
\usepackage{tikz}
\usetikzlibrary{shadows}
\usetikzlibrary{shapes.geometric}
\usetikzlibrary{positioning}
\usetikzlibrary{fit}
\usetikzlibrary{backgrounds}
\usetikzlibrary{arrows.meta}
\usepackage{placeins}  %

\usepackage{newfloat}
\usepackage{listings}

\newcommand{\defined}[1]{\textbf{#1}}

\usepackage{pifont}
\newcommand{\cmark}{\ding{51}}
\newcommand{\xmark}{\ding{55}}

\newcommand{\ie}{i.\,e.}
\newcommand{\eg}{e.\,g.}

\newcommand{\powerset}[1]{\mathcal{\wp}(#1)}

\newcommand{\tuple}[1]{\ensuremath{\langle #1 \rangle}}

\definecolor{tabblue}{HTML}{1f77b4}
\definecolor{taborange}{HTML}{ff7f0e}
\definecolor{tabgreen}{HTML}{2ca02c}
\definecolor{tabred}{HTML}{d62728}
\definecolor{tabpurple}{HTML}{9467bd}
\definecolor{tabbrown}{HTML}{8c564b}
\definecolor{tabpink}{HTML}{e377c2}
\definecolor{tabgray}{HTML}{7f7f7f}
\definecolor{tabolive}{HTML}{bcbd22}
\definecolor{tabcyan}{HTML}{17becf}

\newcommand{\predicates}{\ensuremath{\mathcal{P}}}

\newcommand{\objects}{\ensuremath{\mathcal{O}}}

\newcommand{\acts}{\ensuremath{\mathcal{A}}}
\newcommand{\init}{\ensuremath{I}}
\newcommand{\goal}{\ensuremath{G}}

\newcommand{\pre}{\ensuremath{\mathit{pre}}}

\newcommand{\add}{\ensuremath{\mathit{add}}}
\newcommand{\del}{\ensuremath{\mathit{del}}}

\newcommand{\apply}[1]{\ensuremath{\llbracket#1\rrbracket}}

\newcommand{\goalref}{\ensuremath{G^{\text{\textup{ref}}}}}
\newcommand{\goalenf}{\ensuremath{G^{\text{\textup{enf}}}}}
\newcommand{\goaltrue}{\ensuremath{G^{\text{\textup{true}}}}}
\newcommand{\goalfalse}{\ensuremath{G^{\text{\textup{false}}}}}

\newcommand{\allgoals}{\ensuremath{\bigcup \goalref_i}}
\newcommand{\allgoalsref}{\ensuremath{\goalref_{all}}}

\newcommand{\trace}{\ensuremath{\sigma}}

\newcommand{\task}{\ensuremath{\tau}}

\newcommand{\plan}{\ensuremath{\pi}}
\newcommand{\plans}{\ensuremath{\Pi}}

\newcommand{\explanations}{\ensuremath{\mathcal{E}}}
\newcommand{\explanation}{\ensuremath{E}}
\newcommand{\args}{\ensuremath{\textit{args}}}

\newcommand{\minsubset}{\min_\subseteq}

\newcommand{\iterstep}{\ensuremath{\delta}}
\newcommand{\allitersteps}{\ensuremath{\Delta}}

\newcommand{\questionf}{\ensuremath{Q}}
\newcommand{\questionNL}{\ensuremath{Q_{\text{NL}}}}

\newcommand{\explanationnNL}{\ensuremath{E_{\text{NL}}}}
\newcommand{\responseNL}{\ensuremath{R_{\text{NL}}}}
\newcommand{\goalNL}{\ensuremath{G_{\text{NL}}}}
\newcommand{\flanguage}[1]{\ensuremath{{\mathcal L}(#1)}}
\newcommand{\flanguageb}[1]{\ensuremath{{\mathcal L}_{#1}}}

\newcommand{\questiontypes}{\ensuremath{{M_Q}}}
\newcommand{\questiontype}{\ensuremath{{\mu_Q}}}

\newcommand{\goalTrans}{\ensuremath{T_G}}
\newcommand{\questionTrans}{\ensuremath{T_Q}}

\newcommand{\questionTopicTrans}{\ensuremath{T_{T}}}
\newcommand{\explanationTrans}{\ensuremath{T_\explanations}}
\newcommand{\questionSugg}{\ensuremath{S_{Q}}}

\newcommand{\allmugs}{\ensuremath{{\mathcal G}^{\textup{MUS}}}}

\newcommand{\allmcs}{\ensuremath{{\mathcal G}^{\textup{MCS}}}}

\newcommand{\HIT}{\textit{HIT}}

\newcommand{\explanationframework}{\ensuremath{\text{EF}_{\text{CC}}}}

\newcommand{\LTLf}{LTL\ensuremath{_\text{f}}}
\newcommand{\LTLnot}{\neg}
\newcommand{\LTLalways}{\square}

\newcommand{\LTLuntil}{\text{ U }}

\newcommand{\uswhynot}{\texttt{US-why}}
\newcommand{\ushow}{\texttt{US-how}}
\newcommand{\swhynot}{\texttt{S-why-not}}
\newcommand{\swhatif}{\texttt{S-what-if}}
\newcommand{\scan}{\texttt{S-can}}
\newcommand{\showqtype}{\texttt{S-how}}

\newcommand{\xfquerySkipGT}{\texttt{EFQUERY-noGT}}
\newcommand{\xfqueryWithGT}{\texttt{EFQUERY-GT}}
\newcommand{\direct}{\texttt{DIRECT}}
\newcommand{\followup}{\texttt{FOLLOW-UP}}

\newcommand{\templateBased}{\texttt{G}$^\texttt{TPL}$}
\newcommand{\LLMBased}{\texttt{G}$^{\texttt{LLM}}$}

\renewcommand{\orcidID}[1]{}  %

\begin{document}
\title{Exploring Plan Space through Conversation: \\ An Agentic Framework for LLM-Mediated Explanations in Planning}
\titlerunning{Exploring Plan Space through Conversation}
\author{Guilhem Fouilhé\inst{1}\orcidID{0000-1111-2222-3333} \and
  Rebecca Eifler\inst{2}\orcidID{1111-2222-3333-4444} \and
  Antonin Poché\inst{1,3}\orcidID{2222--3333-4444-5555} \and \\
  Sylvie Thiébaux\inst{2,5}\orcidID{3333--4444-5555-6666} \and
  Nicholas Asher\inst{1,4}\orcidID{4444--5555-6666-7777} }
\authorrunning{G. Fouilhé et al.}
\institute{IRIT, Toulouse, France\and
  LAAS-CNRS, Toulouse, France\and
  IRT Saint Exupery, Toulouse, France\and
  CNRS, Toulouse, France\and
  Australian National University, Canberra, Australia}
\maketitle              %
\begin{abstract}
  When automating plan generation for a real-world sequential decision problem, the goal is often not to replace the human planner, but to facilitate an iterative reasoning and elicitation process, where the human's role is to guide the AI planner according to their preferences and expertise. In this context, explanations that respond to users' questions are crucial to improve their understanding of potential solutions and increase their trust in the system. To enable natural interaction with such a system, we present a multi-agent Large Language Model (LLM) architecture that is agnostic to the explanation framework and enables user- and context-dependent interactive explanations.  We also describe an instantiation of this framework for goal-conflict explanations, which we use to conduct a user study comparing the LLM-powered interaction with a baseline template-based explanation interface.

  \keywords{Conversational Explanations  \and AI Planning \and LLM Agents}
\end{abstract}

\section{Introduction}
\label{sec:introduction}

One can think of planning as the task of finding a plan that satisfies a set of properties, but also as the iterative process that starts before the goals, objectives, and preferences are fully defined \cite{smith:aaai-12}, and ends when a plan is found that is satisfactory for all parties involved. From this perspective, explanations serve the purpose of accelerating the convergence of preferences elicitation by humans. Explanations are by nature interactive; an explanation is something that one person (the explainer) does---it's a speech act \cite{searle:1979}--- for the sake of another (the explainee).
But while interactive planning has become a relatively well-accepted paradigm, the crucial role of interactive \textit{explanations} in that process has been less widely investigated.

The interactive process of finding suitable goals and constraints can go off track when agent A proposes a property that agent B rejects.
Assuming this process is cooperative, A will have proposed P on the assumption that this would help B in finding a suitable plan.  So A will be in situation of a conceptual predicament, which on the pragmatic view of explanations we espouse, is a precondition for B's providing an explanation of their action \cite{bromberger:1962,achinstein:1980}.  The point of B's explanation is to clear up A's conceptual problem so that the interactive process of finding goals and constraints can resume and continue.
Explanations thus are ideally designed to get the interactive process back on track.

However, fixing a conceptual problem may not be a one-shot deal. A conceptual problem is typically the result of multiple factors.  In the scenario above, B's explanation for why P is unsatisfactory may rely on assumptions Q, R, ... that A may question or find incompatible with their view of the situation.  Thus, an explanation answering the question {\em why not P?} may give rise further explanations clarifying why Q or R. it may take several interactions, several related explanations, to resolve A's conceptual problem.

Such interaction is beneficial; previous work in psychology and human-machine interaction \cite{liao2020questioning,dazeley2021levels,lakkaraju2022rethinking,zhang2024may} provides evidence that interactive explanations are more interpretable and effective than static ones \cite{miller:ai-19}.

\begin{figure}[!tp]
  \centering
  \includegraphics[scale=0.9]{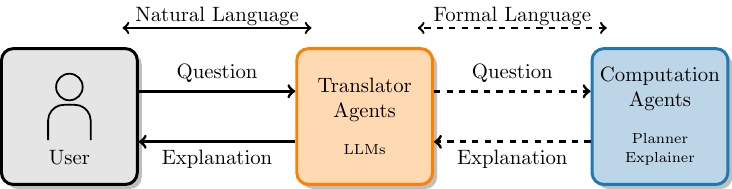}
  \caption{
    High-level overview of our approach to interactive planning with explanations.
    Translator agents based on LLMs translate the
    user input into the formal language required by the computation agents.
  }
  \label{fig:overview-architecture}
\end{figure}

Explanations between humans are interactive; hence, to provide appropriate explanations, we need conversational capacities that support an initial request for an explanation, as well as follow-up questions and remarks, in a conversationally fluid manner. State-of-the-art LLMs address this need. However,
LLMs cannot do everything.  They cannot handle the full interactive planning task. \cite{valmeekam:etal:2023,corrêa20252025planningperformancefrontier} provide empirical evidence that
LLMs cannot reliably perform complex planning tasks independently, despite remarkable progress with reasoning models, and \cite{peng:etal:2024} gives theoretical
reasons to suspect models using current transformer architectures never will.
Our work extends the view of \cite{kambhampati2024llms} that LLMs can nevertheless serve valuable supporting roles in a hybrid planning system.
Figure~\ref{fig:overview-architecture},
shows how we integrate LLMs into the planning process
while delegating the core computational planning tasks to specialized algorithms
designed for that purpose.

To enable
reliable natural interactions between users and planning systems, we need to
accommodate different ways in which a user formulates their question.
To this end, we leverage LLMs in several distinct roles: as a classifier of the
question type, as a translator of the question topic, and as a question suggester
that generates example relevant questions to help users articulate their information needs.
These LLM agents can then query an explanation system that can produce a formal explanation. To deliver this explanation to the user, we use an LLM as an explanation translator. It converts this formal explanation into a natural language response that can serve as the basis for an extended dialogue, in which the original explanation is refined to meet the user's needs.

Our evaluation, based on a user study with lay participants, focuses on human perception and objective improvements at helping individuals to achieve a planning task.

Our contributions are the following :

\begin{itemize}[left=0pt]
  \item We formalize a framework for iterative planning with natural language questions and explanations
        (Section~\ref{sec:iterative_planning_with_explanations}).
  \item We instantiate this framework with goal-conflict explanations \cite{eifler:etal:aaai-20} and extend the diversity of questions that can be handled with the latter (Section~\ref{sec:goal-onflict-explanation}).
  \item We introduce an LLM-based implementation of our framework for natural language interaction (Section~\ref{sec:translator-llm-implementation}).
  \item We evaluate and validate our framework in a user study, comparing our LLM-based interface with a template-based baseline (Section~\ref{sec:evaluation}).
\end{itemize}

\section{Background}
\label{sec:background}

\subsection{Conversational Explanations}

Social science and XAI research suggest explanations are most effective when delivered conversationally or interactively. %
Other methods have been proposed for interactive explanations \cite{nguyen2023black,shen2023convxai,slack2023explaining,state2023reason,feldhus2023interrolang,OptiChat2025}; they employ a similar high-level architecture, as shown in Figure~\ref{fig:overview-architecture}, but none of them are specifically designed for planning systems.

\subsection{Multi-Agent LLM Approaches}

Multi-agent LLM approaches involve multiple LLMs working collaboratively to solve complex tasks \cite{guo2024large}. These systems typically leverage individual LLM strengths by assigning specific roles to each agent, allowing for specialization and improved performance through in-context learning tailored to particular functions.

While a single LLM can be fine-tuned for complex planning tasks, this faces key challenges: insufficient task-specific training data and limited generalization beyond the training distribution.
To address these limitations, we propose decomposing complex tasks into specialized subtasks handled by different LLM agents working in concert. This architecture leverages LLMs' natural strengths in language understanding and translation while mitigating their weaknesses in complex reasoning tasks. Such approaches often show significant performance gains over single-agent approaches, resulting in a  "greater-than-the-sum-of-its-parts" system \cite{riedlEmergentCoordinationMultiAgent2025,Chen2024May}.

\subsection{Planning Formalism}

We define a lifted planning task as a tuple $\task =
    \tuple{\predicates,\allowbreak\objects,\allowbreak\acts,\allowbreak\init,\allowbreak\goal}$
where $\predicates$ is a finite set of first-order predicates, $\objects$
a set of individual constants or names for objects, $\acts$ a set of action types or schemas, $\init$ an initial state and $\goal$ a set of goals, which we specify below.
$P$ is a ground predicate or \defined{atom} if all variables have been replaced
by individual constants or names objects.
The \defined{goals} $\goal$ is a set of atoms.
A \defined{state} is a set of atoms; atoms not in the set are assumed to be
false in the state. $\init$ is the \defined{initial state}.
Each action schema $A \in \acts$ has a list $X_A$ of parameter variables and a
\defined{precondition} $\pre_A$, an \defined{add list} $\add_A$ and
a \defined{delete list} $\del_A$ which are sets of predicates from $\predicates$
where all variables are replaced by an element in $X_A \cup \objects$. The add and delete lists are used to define the effects of an action.
We obtain a (ground) action $a$ from an action schema $A$ by replacing all
variables $X_A$ in $\pre_A$, $\add_A$, and $\del_A$ with an object from $\objects$.
A action $a$ is \defined{applicable} in a state $s$ if $\pre_a \subseteq s$ and
$\acts(s)$ denotes the set of all applicable actions in $s$.
Applying action $a$ in state $s$,
results in the state $s\apply{a} = (s \setminus \del_a) \cup \add_a $.
The state resulting from an iteratively applicable sequence of actions
$\plan \!=\!\tuple{a_1, \ldots, a_n}$ is denoted by $s\apply{\plan}$.
A \defined{plan} is an action sequence $\plan$ such that
$\goal \subseteq I\apply{\plan}$.
A task $\task$ is \defined{solvable} if a plan exists.
By $\plans(\task)$ we denote the set of all plans for
task $\task$ and by $\task(G')$ we denote the task
$\task' \!= \!\tuple{\predicates,\allowbreak\objects,\allowbreak\acts,\allowbreak\init,\allowbreak\goal'}$

\subsection{Goal-Conflict Explanations}
In the following, we consider a setting similar to oversubscription planning \cite{smith:icaps-04,domshlak:mirkis:jair-15}, where not all goals can be satisfied, for example, due to insufficient resources. Within this framework, the objective is to find a subset of goals that maximizes utility; here, we examine the conflicts between a set of reference goals $\goalref$.
For a task $\task$, these conflicts and possible resolutions are given by \defined{minimal unsolvable subsets} (MUS) \cite{eifler:etal:aaai-20} and \defined{minimal correction sets} (MCS) respectively.

\begin{definition}[Minimal Unsolvable Subset (MUS)]
    A set of goals $C \subseteq \goalref$ is a \textbf{MUS} if $\task(C)$ is unsolvable but for all $G \subset C$, $\task(G)$ is solvable.
\end{definition}

\begin{definition}[Minimal Correction Set (MCS)]
    A set of goals $R \subseteq \goalref$ is a \textbf{MCS} if $\task(\goalref \setminus R)$ is solvable but for all $G \subset R$, $\task(\goalref \setminus G)$ is unsolvable.
\end{definition}

By $\allmugs(\task, \goalref)$ and $\allmcs(\task, \goalref)$ we denote the set of all MUS and MCS for task $\task$ with respect to the reference goals $\goalref$. Both sets can be exponentially large when $|\goalref|$ increases. The following relation holds between MUS and MCS over the same set of goals $G$: $\allmcs(\task, G) = \HIT(\allmugs(\task, G))$, where $\HIT(\mathcal{S})$ is the set of all minimal hitting sets of the sets in $\mathcal{S}$. For algorithms to compute $\allmugs(\task, G)$ we refer to \cite{eifler:etal:aaai-20,eifler:etal:ijcai-20}.

\subsection{Temporal Goals}
\cite{eifler:etal:ijcai-20} introduced algorithms that employ a compilation approach to compute conflicts over goal facts and temporal plan properties. Such properties are described using Linear Temporal Logic on finite traces (\LTLf) over atoms \cite{giacomo:etal:aaai-14}. Our framework also supports temporal goals and bridges a gap with previous work by enabling natural language-based temporal goal creation. Translating natural language to LTL or \LTLf\ has been explored with different approaches \cite{brunello:etal:time19}. More recently, prompting-based LLM methods implement tools such as \texttt{NL2LTL} \cite{fuggitti:etal:icaps23}, a template-based classifier, \texttt{Lang2LTL} \cite{liu:etal:LRWCoRl22}, which works without templates and only provides the available literals to the LLM, and \texttt{nl2spec} \cite{cosler:etal:CAV23}, which addresses sub-formulas iteratively to counteract ambiguities.

$\flanguage{\task}$ is the set of well-formed \LTLf\ formulas of task $\task$.

\subsection{Running Example} \label{sec:running_example}

To illustrate the different concepts of our framework, we introduce a small running example based on the Parents' Afternoon task used in the user study detailed in Section~\ref{sec:evaluation}. We introduce three goals with conflicting priorities: shopping ($S$), cooking dinner ($C$), and taking the kids to their sports match ($M$). The match and cooking are both evening activities, leading to a conflict between the two goals. Formally, the reference goals are $G_{ex}^{ref} = \{S, C, M\}$, there is one MUS: $\allmugs_{ex} = \bigl\{\{C, M\}\bigr\}$, and two MCS: $\allmcs_{ex} = \bigl\{\{C\},\{M\}\bigr\}$. Figure~\ref{fig:running_example} below illustrates the example.

\begin{figure}[htbp]
    \centering
    \begin{tikzpicture}[
            node distance=4cm,
            goal/.style={draw, circle, minimum size=2cm, align=center, fill=blue!10, thick, font=\sffamily},
            conflict/.style={<->, red, ultra thick, dashed}
        ]

        \node[goal, xshift=1cm] (S) {$S$ \\ \scriptsize Shopping};

        \node[below=1.3cm of S, align=center, font=\sffamily\color{gray}] (indep_s) {$S$ is not involved \\ in any conflict \\ (no tradeoff needed)};
        \draw[->, thick, dashed, draw=gray] (S) -- (indep_s);

        \node[goal, right=2.5cm of S, yshift=1.8cm] (C) {$C$ \\ \scriptsize Cooking};
        \node[goal, right=2.5cm of S, yshift=-1.8cm] (M) {$M$ \\ \scriptsize Match};

        \begin{scope}[on background layer]
            \node[
                draw, dashed, fill=orange!10, rounded corners,
                inner xsep=1.5cm, inner ysep=0.5cm,
                fit=(C) (M),
                label={[font=\sffamily\color{orange!80!black}]above:Evening time (Conflict)}
            ] (evening_box) {};

            \node[
                draw, densely dotted, gray, thick, rounded corners,
                inner xsep=0.5cm, inner ysep=1cm,
                fit=(S) (evening_box),
                label={[font=\sffamily\bfseries\color{gray}]south:Reference Goals: $G_{ex}^{ref} = \{S, C, M\}$}
            ] (ref_box) {};
        \end{scope}

        \draw[conflict] (C) -- node[right, text=red, font=\sffamily\bfseries, align=left, xshift=0.2cm, draw, fill=red!5, rounded corners] {$\{C,M\} \in $ MUS: The task is unsolvable as long \\ as both $C$ and $M$ are enforced} (M);

        \node[right=2.2cm of C, align=left, font=\sffamily, draw, fill=green!5, rounded corners] (mcs_c) {Removing $C$ makes \\ the task solvable. \\ $\rightarrow$ \textbf{MCS}: $\{C\}$};
        \draw[->, thick, dashed, draw=gray] (C) -- node[above, text=red] {\xmark} (mcs_c);

        \node[right=2.2cm of M, align=left, font=\sffamily, draw, fill=green!5, rounded corners] (mcs_m) {Removing $M$ makes \\ the task solvable. \\ $\rightarrow$ \textbf{MCS}: $\{M\}$};
        \draw[->, thick, dashed, draw=gray] (M) -- node[above, text=red] {\xmark} (mcs_m);

    \end{tikzpicture}
    \caption{Illustration of the running example: Reference goals ($G_{ex}^{ref}$), the Minimal Unsatisfiable Subset ($\allmugs_{ex}$) causing a time conflict during the evening, and the Minimal Correction Subsets ($\allmcs_{ex}$) that resolve the conflict by removing either goal.}
    \label{fig:running_example}
\end{figure}

\section{Iterative Planning with Natural Language}
\label{sec:iterative_planning_with_explanations}

In this section, we provide an abstract description of iterative planning with conversation. We define a generic framework for iterative planning with explanations, focusing on user interaction points.

In our framework, users refine their preferences through iterative steps (Definition~\ref{def:iter-step}). Each step is associated with a solution plan if it exists. In a step, the user can ask questions (\ref{def:question}) to refine or modify the current plan. As seen earlier, questions have both a type and a topic. In the rest of this paper, we will consider \emph{goals} as question topics. The question is translated (\ref{def:question-translator}) to a particular type and linked to a formal goal (\ref{def:goal-translator}), which may or may not already be known to the system. Once the system has a formal goal and a question type, it computes a formal explanation and translates it (\ref{def:explanation-translator}) into natural language for the user.

More formally, the iterative process of determining a final plan is divided into individual steps. Each step $\iterstep_i$ represents one snapshot of the user's exploration of the plan space, defined by a set of reference goals $\goalref_i$, a set of enforced goals $\goalenf_i \subseteq \goalref_i$, and a sample plan $\plan_i$ satisfying $\goalenf_i$:

\begin{definition}[Iteration Step]\label{def:iter-step}
   Given a planning task $\task$, an \defined{iteration step} is a tuple
   $\iterstep = \tuple{\goalref, \goalenf, \plan}$,
   where $\goalref$ is a set of (temporal) goals for $\task$,
   $\goalenf \subseteq \goalref$,
   and $\plan \in \plans(\task(\goalenf))$ if $\task(\goalenf)$ is solvable and
   $\plan = \epsilon$ otherwise.
\end{definition}

Based on previous plans $\plan_{i}$, the user defines the set of reference goals $\goalref_{i+1}$ for the next step, representing the goals and preferences they want the plan to satisfy. The user also selects a subset of goals $\goalenf_{i+1} \subseteq \goalref_{i+1}$ that must be satisfied by the sample plan $\plan_{i+1}$ in the next iteration.

\paragraph{Goals}  The reference goals must be defined in a language the planner understands.  For this we use \LTLf. However, this language is not suitable for a lay person as an input language. Thus, we require a goal translator.

\begin{definition}[Goal Translator] \label{def:goal-translator}
   Given a task $\task$ with well-formed temporal goals $\flanguage{\task}$,
   a \defined{goal translator} is a function
   $\goalTrans: \textit{NL} \mapsto \flanguage{\task} \cup \epsilon$,
   that maps a natural language input to a goal $\phi \in \flanguage{\task}$ and
   to $\epsilon$ if the natural language description does not describe a
   goal represented by any formula in \flanguage{\task}.
\end{definition}

In our running example of Section \ref{sec:running_example}, the goal translator allows us to create a new goal, converting e.g. "never bring kids shopping" to "$\LTLalways \LTLnot at(kid,store)$". This new goal adds a constraint to the planner and subject to questions.

A more advanced goal translator could include a feedback loop with the user to recover from misunderstandings or alert them about similar goals already being considered. These features rely on the interaction context, which captures previous interactions.

Explanations help the user decide which goals to enforce in the next iteration. They can clarify the model \cite{sreedharan:etal:ai21}, identify trade-offs between plan quality measures \cite{krarup:etal:icaps24}, or improve understanding of dependencies between the goals, which is the focus of this paper. Explanations are provided as answers to specific user questions. Depending on whether the enforced goals $\goalenf_i$
are (i) satisfiable ---a sample plan %
exists--- or (ii) unsolvable, the user's questions will differ: given (i), they may ask how the sample plan solves the task; given (ii), they will ask why $\goalenf_i$ cannot be satisfied.

\paragraph{Questions} Formally, we define a question as follows.

\begin{definition}[Question] \label{def:question}
   Given a task $\task$ and a set of question types $\questiontypes$,
   a \defined{question} is a tuple $\questionf = \tuple{\questiontype, \textit{args}}$
   where $\questiontype \in \questiontypes$ is the question type and
   $\textit{args} \subseteq \powerset{\flanguage{\task}}$
   (the powerset of $\flanguage{\task}$) are the question arguments.
\end{definition}

The question types $M_Q$ depend on the explanation framework. In Section~\ref{sec:goal-onflict-explanation} we introduce the types our framework supports. An example is $\questiontype = \swhynot$, \ie\
``Why is $g$ not satisfied by plan $\plan$?'', which requires one argument. Roughly, each question word in natural language, \eg\ \emph{who, what, why, how} maps to a different type of question.

This is the second point of interaction. Again, we require a translator to allow the user to ask their question in natural language, which is then translated into a formal question.

\begin{definition}[Question Translator] \label{def:question-translator}
   Given a task $\task$, and a set of question types $M_Q$, a \defined{question translator} is a function $\questionTrans: \textit{NL} \mapsto (M_Q \times \powerset{\flanguage{\task}}) \cup \epsilon$, that maps a natural language expression to a question type and its arguments and to $\epsilon$ if no matching question type exists.
\end{definition}

In our running example introduced in Section \ref{sec:running_example}, the question translator would map "Do I have the time to buy shoes" to "$\scan \{S\}$". Simultaneously classifying the question type and mapping the arguments to a known goal. See \ref{sec:goal-onflict-explanation} for question types details.

Question translation can be divided into two steps. First, the question type is identified. Then, the question topic, \ie\ the arguments, are translated. The goal translator can perform this last step, as the translation tasks are the same.

A context-dependent question translator with access to previous questions enables follow-ups with implicit references. For example, in ``Why can I not visit Alice?'' and ``Can you enforce it?'', the visited location or even the entire question argument depends on the context.

\paragraph{Selecting Questions}
In a pilot user study, lay users often struggled to formulate questions. As a result, they made less use of the explanation interface than intended and produced only a narrow range of questions. This suggests that a crucial task for an explanation interface is to support users in generating meaningful, diverse inquiries,
for instance, by suggesting questions. While allowing users to ask any natural language question is a desirable feature of a conversational interface, it can increase cognitive load \cite{Nguyen2022Mar}. The Question Suggester provides a sample of relevant questions to mitigate this.%

\begin{definition}[Question Suggester] \label{def:question-suggester}
   A \defined{Question Suggester} is a function $\questionSugg: \iterstep \mapsto \textit{NL}^k$ that maps a given iteration step to $k$ relevant natural language questions.
\end{definition}

\paragraph{Explanations} The explanation framework computes a set of formal explanations $\explanations(\questionf)$ based on the translated question $\questionf = \tuple{\questiontype, \textit{args}}$ and additional required data, such as the planning task $\task$ and the current iteration step $\iterstep_i$. We do not place any restrictions on the exact format or language of an explanation $\explanation \in \textit{explanations}(\questionf)$, and will refer to the language as $\flanguageb{\explanations}$. However, we assume that each explanation $\explanation \in \explanations(\questionf)$ is sufficient in itself to answer the question. $\explanations(\questionf)$ is regarded as a selection of possible explanations.

These explanations must be communicated to the user, and thus, we again need a translator function.

\begin{definition}[Explanation Translator] \label{def:explanation-translator}
   An \defined{explanation translator} is a function $\explanationTrans: \powerset{\flanguage{\explanations}} \mapsto \textit{NL}$, that maps a set of formal explanations to a natural language explanation.
\end{definition}

As an example of goal-conflict explanations, following the running example: suppose the current iteration step achieves goal $M$. The question “How can I find time to cook dinner?”, translated as $\tuple{\showqtype, \{C\}}$, yields $\explanations(\tuple{\showqtype, \{C\}}) = \{\{M\}\}$. This means that $C$ is achievable only if at least $M$ is dropped in the next iteration step. Hence, an explanation translator would respond: “You’d need to cancel the sports match.”. How this explanation is computed is detailed in the next section.

This is a simple version of an explanation translator that does not provide user- or context-dependent translation.
User-dependent translation is crucial for customizing the vocabulary to the user and providing an answer with the expected level of detail. Incorporating the interaction context can have several benefits, from the possibility of asking follow-up questions, which naturally increases interactivity, to the inclusion of previous interactions in the selection and summarization of explanations.

\paragraph{Selecting Explanations}
The explanation framework provides a set of explanations, each of which addresses the question $\questionf$ independently. It may be preferable to convey multiple explanations $\hat{\explanations} \subseteq \explanations(\questionf)$, in cases where the effect has multiple sufficient causes, neither of which is necessary (over-determination). In cases without over-determination, some causes or properties will not be as relevant as others or may not fit into the interaction context. In line with the insight from social science \cite{miller:ai-19} that humans tend to select small, relevant explanations given the context, using only a subset of explanations can be advantageous.

Often the size of the explanations is chosen as selection criteria \cite{chakraborti:etal:ijcai17}. \cite{junker:aaai04} selects \emph{preferred} explanations based on an upstream or interleaved process to collect a preference ranking over the goals.

\paragraph{Summarizing Explanations}

In case the model returns a large set of explanations and these are at different levels and expressing their logical
relation results in an overly lengthy response, it is often simpler to provide
a summary of the explanations $\hat{\explanations}$.
In line with how humans give explanations \cite{miller:ai-19}, such a
summary should convey only relevant information as
effectively as possible in the given context.
As for the explanation selection, the user questions can provide an indication of
the level of detail expected.
Also, the interaction context can be used to enable or facilitate the summarization
by referring to previously addressed questions or explanations.

Selecting the correct abstraction level for the explanations
\cite{sreedharan:etal:ijcai19}, or only communicating via a predefined
vocabulary \cite{vasileiou:etal:ecai23} also leads to a summarization
in the sense of leaving out details that are not required or known.

\section{Richer Goal Conflict Explanations}
\label{sec:goal-onflict-explanation}

We address the following question types, allowing a rich user interaction (see Table~\ref{tab:question-types}).
This significantly extends the framework by \cite{eifler:etal:aaai-20} which focused on \swhynot\ questions.

\begin{table}[h]
      \centering
      \begin{tabular}{lp{4cm}c}
            \toprule
            \textbf{Type} & \textbf{Formulation}                    & \textbf{New} \\
            \midrule
            \uswhynot     & Why is the task unsolvable?             & \cmark       \\
            \ushow        & How can I make the task solvable?       & \cmark       \\
            \swhynot      & Why are $\textit{\args}$ not satisfied? & \xmark       \\
            \swhatif      & What if I enforce $\textit{\args}$?     & \cmark       \\
            \scan         & Can $\textit{\args}$ be enforced?       & \cmark       \\
            \showqtype    & How can $\textit{\args}$ be achieved?   & \cmark       \\
            \bottomrule
      \end{tabular}
      \caption{Question types supported by \explanationframework}
      \label{tab:question-types}
\end{table}

These question types were collected during an interview conducted at a manufacturing company
with the end-users of a future explainable planning system.
We call this \textit{Explanation Framework with Conflicts and Corrections} (\explanationframework), in line with the fact that all answers are based on either the minimal conflicts (MUS) or minimal corrections (MCS) of the reference goals $\goalref$ and the arguments of the question.

In the following definitions, we assume the user asks questions about goals already included in $\goalref$. We list all supported question types and the corresponding formal explanations (prefaced by bullets). To clarify the meaning of the question and answer, we include one possible natural language version of both. As an input, we require the task $\task$, the iteration step $\iterstep = \tuple{\goalref, \goalenf, \plan}$ and the question $\questionf = \tuple{\questiontype, \textit{\args}}$. The produced explanations $\explanations(\questionf)$ are subsets of $\goalref$, thus explanations' formal language is $\flanguageb{\explanations} = \powerset{\flanguage{\task}}$.

\subsection{Unsolvable}
If $\task(\goalenf)$ is unsolvable we support the following two question types:

\begin{itemize}[left=0pt]
      \item \uswhynot: ``Why is the task unsolvable?''
            \begin{align*}
                  \explanations(\tuple{\uswhynot, \emptyset}) =\allmugs(\task,\goalenf)
            \end{align*}
            Sample explanation: ``The task is unsolvable because it is not possible to satisfy any of the conflicts in $\explanations(\tuple{\uswhynot, \emptyset})$.''

      \item \ushow:  ``How can I make the task solvable?''
            \begin{align*}
                  \explanations(\tuple{\ushow, \emptyset}) = \allmcs(\task, \goalenf)
            \end{align*}
            Sample explanation: ``To make the task solvable you have to forego one of the goal sets in $\explanations(\tuple{\ushow, \emptyset})$.''
\end{itemize}

\noindent
\subsection{Solvable} If $\task(\goalenf)$ is solvable, we support question types referring to goals not satisfied by the sample plan $\plan$. By $\goaltrue(\plan) =  \{\phi \in \goalref \mid \trace(\plan, I) \models \phi\}$ we denote the goals satisfied by $\plan$, and by $\goalfalse(\plan) = \goalref \setminus \goaltrue(\plan)$ the goals not satisfied by $\plan$. For all the following question types, the arguments must not be satisfied by the current plan, i.e. $\args \subseteq \goalfalse(\plan)$. Answers to question types $\{\swhynot,\swhatif,\scan\}$ use the same information, but are phrased differently.
\begin{align*}
      \explanations(\tuple{\questiontype, \args}) & = \minsubset(\{C \setminus \args \mid
      C \in \allmugs(\task, \goalref), C \subseteq \goaltrue(\plan) \cup \args \})
\end{align*}
\noindent
where $\minsubset(S)$ filters the sets in $S$ that are subset-minimal.

\begin{itemize}[left=0pt]
      \item \swhynot: ``Why are $\args$ not satisfied?''

            Sample explanations:
            \begin{itemize}[left=5pt]
                  \item If $\explanations(\tuple{\swhynot, \args}) = \emptyset$:
                        ``$\args$ can be satisfied without foregoing any of the already satisfied goals.''
                  \item If $\emptyset \in \explanations(\tuple{\swhynot, \args}) $:
                        ``The goals in $\args$ cannot be satisfied together.''
                  \item Otherwise:
                        ``There is a conflict between $\args$ and all the goal subsets in $\explanations(\tuple{\swhynot, \args})$.''
            \end{itemize}

      \item \swhatif: ``What happens if we enforce $\args$?''

            Sample explanations:
            \begin{itemize}[left=5pt]
                  \item If $\explanations(\tuple{\swhatif, \args})  = \emptyset$:
                        ``$\args$ can be satisfied without foregoing any goal satisfied by the plan.''
                  \item If $\emptyset \in \explanations(\tuple{\swhatif, \args})$:
                        ``Then the problem would be unsolvable.''
                  \item Otherwise: ``You could no longer satisfy any of the goal sets in
                        $\explanations(\tuple{\swhatif, \args})$.''
            \end{itemize}

      \item \scan: ``Can $\args$ be satisfied?''

            In this yes/no question type, it is implicit that no currently satisfied goal should be given up. This is supported by human planners' sample responses, which never considered foregoing goals in $\goaltrue(\plan)$.

            Sample explanations:
            \begin{itemize}[left=5pt]
                  \item If $\explanations(\tuple{\scan, \args})  = \emptyset$: ``$\args$ can be satisfied.''
                  \item Otherwise: ``It is not possible.''
            \end{itemize}

      \item \showqtype: ``How can $\args$ be satisfied?''
            \begin{align*}
                  \explanations(\tuple{\showqtype, \args}) & = \minsubset(\{C \setminus \goalfalse(\plan) \mid
                  C \in \allmcs(\task, \goalref), C \cap \args = \emptyset\})
            \end{align*}
            Sample explanations:
            \begin{itemize}[left=5pt]
                  \item If $\emptyset \in \explanations(\tuple{\showqtype, \args})$:
                        ``$\args$ can be satisfied without foregoing any goals satisfied by the plan.
                  \item If $\explanations(\tuple{\showqtype, \args}) = \emptyset$:
                        ``It is not possible.''
                  \item Otherwise: ``You have to forego one of the goal sets in
                        $\explanations(\tuple{\showqtype, \args})$.''
            \end{itemize}
\end{itemize}

\noindent
The numbers of MUSs and MCSs can be exponential in the number of goals. For any question type $\questiontypes$, however, a single goal subset $\explanation \in \explanations(\questionf)$ suffices. Since some conflicts are more relevant or some corrections easier to forego, selecting $\hat{\explanations} \subseteq \explanations(\questionf)$ and summarizing $\hat{\explanations}$ is desirable.

\section{LLMs-based Interactive Framework}
\label{sec:translator-llm-implementation}
\begin{figure*}[t]
    \centering
    \newcommand{\userd}[3]{
    \begin{scope}[shift={(#1,#2)}, scale=0.6, every node/.append style={transform shape}]
        \node[draw, circle, line width=0.3mm, minimum size=5mm] (h) at (0,0) {};
        \draw[line width=0.3mm,  rounded corners=0.1cm]
        (-0.4,-1) --
        (-0.4,-0.4) --
        (0,-0.3) --
        (0.4,-0.4) --
        (0.4,-1);
        \node[] (#3) at (0.3,-0.5) {};
    \end{scope}
}

\newcommand{\messagestyle}{fill=tabred!30, draw=tabred!50, line width=0.2mm}

\begin{tikzpicture}[scale=0.82, every node/.append style={transform shape}]

    \scriptsize

    \begin{scope}[shift={(0,0)}]
        \node[
            anchor=center,
            draw,
            fill=black!10,
            line width=0.5mm,
            minimum width=1.5cm,
            minimum height=1.5cm,
            rounded corners=0.2cm,
            align=center,
            drop shadow,
        ]
        (user) at (0,0)
        {};
        \userd{0}{0.4}{u}
        \node[] (us) at (0,-0.4) {User};
    \end{scope}

    \node[
        anchor=center,
        draw=orange,
        fill=orange!30,
        line width=0.5mm,
        minimum width=0.6cm,
        minimum height=0.6cm,
        rounded corners=0.2cm,
        align=center,
        drop shadow,
    ]
    (QS) at (2,1.25)
    {
        Question\\
        Suggester\\
        \questionSugg
    };

    \node[
        anchor=center,
        draw=orange,
        fill=orange!30,
        line width=0.5mm,
        minimum width=1.5cm,
        minimum height=1.5cm,
        rounded corners=0.2cm,
        align=center,
        drop shadow,
    ]
    (QC) at (4,0)
    {
        Question\\
        Translator\\
        \questionTrans
    };

    \node[
        anchor=center,
        draw=orange,
        fill=orange!30,
        line width=0.5mm,
        minimum width=1.5cm,
        minimum height=1.5cm,
        align=center,
        rounded corners=0.2cm,
        drop shadow,
    ]
    (GT) at (9,2.3)
    {
        Goal\\
        Translator\\
        \goalTrans
    };

    \node[
        anchor=center,
        draw=orange,
        fill=orange!30,
        line width=0.5mm,
        minimum width=1.5cm,
        minimum height=1.5cm,
        rounded corners=0.2cm,
        align=center,
        drop shadow,
    ]
    (ET) at (9,-2.3)
    {
        Explanation\\
        Translator\\
        \explanationTrans
    };

    \node[
        diamond,
        aspect=1.5,
        anchor=center,
        draw,
        fill=black!10,
        line width=0.5mm,
        minimum width=1.5cm,
        minimum height=1.5cm,
        rounded corners=0.1cm,
        align=center,
        drop shadow,
    ]
    (D) at (9,0)
    {
        Dispatcher\\
        $D(\questiontype)$
    };

    \node[
        anchor=center,
        draw=tabblue,
        fill=tabblue!30,
        line width=0.5mm,
        minimum width=1.5cm,
        minimum height=1.5cm,
        align=center,
        rounded corners=0.2cm,
        drop shadow,
    ]
    (expF) at (16,0)
    {
        Explanation\\
        Framework\\
        \explanationframework
    };

    \draw[->, line width=0.5mm] (QS.west) -- (user.north east);

    \draw[->, line width=0.5mm] (user.east) --
    node[below, align=center, pos=0.5] {Question}
    node[fill=black!5, draw=black!50, line width=0.2mm, above=1mm, align=center, pos=0.5] {$\questionNL$}
    (QC.west);

    \draw[->, line width=0.5mm] (QC.east) --
    node[below, align=center, pos=0.5] {Question Type}
    node[fill=black!5, draw=black!50, line width=0.2mm, above=1mm, align=center, pos=0.5, align=left] {
        $\questionNL, \questiontype$\\
        $\textcolor{tabpurple}{\responseNL}/\textcolor{cyan}{\goalNL}/\textcolor{tabred}{\phi}/\textcolor{tabred}{\emptyset}$
    }
    (D.west);

    \draw[->, line width=0.5mm] (D.north) --
    node[right, align=center, pos=0.08] {\textcolor{cyan}{\xfqueryWithGT}}
    node[left, align=center, pos=0.6] {Question Topic}
    node[fill=black!5, draw=black!50, line width=0.2mm, right=1mm, align=center, pos=0.6] {$\textcolor{cyan}{\goalNL}$}
    (GT.south);

    \draw[->, dashed, line width=0.5mm] (D.east) --
    node[above, align=center, pos=0.15] {\textcolor{tabred}{\xfquerySkipGT}}
    node[below, align=center, pos=0.65] {Question}
    node[solid, fill=black!5, draw=black!50, line width=0.2mm, above=1mm, align=center, pos=0.65]
    {$\tuple{\questiontype, \args = \{\textcolor{tabred}{\phi}\}/\textcolor{tabred}{\emptyset}}$}
    (expF.west);

    \draw[->, line width=0.5mm] (D.south) --
    node[right, align=center, pos=0.1] {\followup}
    node[fill=black!5, draw=black!50, line width=0.2mm, right=1mm, align=center, pos=0.6] {$\questionNL$}
    node[right=8mm, align=center, pos=0.6] {Question}
    (ET.north);

    \draw[->, line width=0.5mm, out=200,in=330] (D.220) to
    node[above, align=center, pos=0.2, rotate=16] {\textcolor{tabpurple}{\direct}}
    node[above, align=center, pos=0.55] {Response}
    node[fill=black!5, draw=black!50, line width=0.2mm, below=1mm, align=center, pos=0.55] {$\textcolor{tabpurple}{\responseNL}$}
    (user.330);

    \draw[->, dashed, line width=0.5mm] (expF.south) --
    (16,-2.3) --
    node[above, align=center, pos=0.5] {Explanations \& prior Information}
    node[solid, fill=black!5, draw=black!50, line width=0.2mm, below=1mm, align=center, pos=0.5]
    {$\explanations^*(\args), \questionNL, \tuple{\questiontype, \args}$}
    (ET.east);

    \draw[->, line width=0.5mm] (ET.west) --
    node[above, align=center, pos=0.7] {Explanation} (0,-2.3)
    node[fill=black!5, draw=black!50, line width=0.2mm, below=1mm, align=center, pos=0.7] {\explanationnNL}
    (0,-2.3) --
    (user.south);

    \draw[->, dashed, line width=0.5mm] (GT.east) --
    node[above, align=center, pos=0.35] {Question}
    node[solid, fill=black!5, draw=black!50, line width=0.2mm, below=1mm, align=center, pos=0.35] {$\tuple{\questiontype, \args=\{\textcolor{cyan}{\phi}\}}$}
    (16,2.3) --
    (expF.north);

    \draw[->, line width=0.5mm, draw=tabgreen] (user.80) --
    (0.15,2.13) --
    node[below=1mm, align=center, pos=0.45] {Goal Description}
    node[fill=black!5, draw=black!50, line width=0.2mm, below=1mm, align=center, pos=0.65] {$\goalNL$}
    (GT.190);

    \draw[->, line width=0.5mm, dashed, draw=tabgreen] (GT.170) --
    node[above=1mm, align=center, pos=0.60] {Goal}
    node[solid, fill=black!5, draw=black!50, line width=0.2mm, above=1mm, align=center, pos=0.45] {$\phi$, goal name}
    (-0.12,2.4) --
    (user.100);

    \begin{scope}[shift={(0.5,-3.5)}, anchor=center]
        \draw[->, line width=0.5mm] (0,0) to (0.5,0);
        \node[anchor=west] (NL) at (0.5,0) {natural language};
        \draw[->, dashed, line width=0.5mm] (2.8,0) to (3.3,0);
        \node[anchor=west] (NL) at (3.3,0) {formal language};

        \node[anchor=west, draw=tabgreen,fill=tabgreen, minimum size=0.5, label=right:{goal creation }] (X) at (5.5,0) {};
        \node[anchor=west, draw,fill=black, minimum size=0.5, label=right:{explanations protocol }] (X) at (7.5,0) {};

        \node[anchor=west, solid, fill=black!5, draw=black!50, line width=0.2mm, minimum width=5mm, minimum height=3mm, label=right:{message}] (X) at (10.6,-0) {};
        \node[anchor=west, solid, fill=taborange!30, draw=taborange, line width=0.5mm, minimum width=3mm, minimum height=3mm, rounded corners=0.05cm, label=right:{LLM Agent}] (X) at (12.4,0) {};
    \end{scope}

\end{tikzpicture}
    \caption{
        Communication protocol between the user, the translators and the explanation framework.
        In addition to the information provided in the messages, each agent has
        access to the planning task and the iteration step.
        For a goal translation (green, top left) the user directly communicates with
        \goalTrans.
        If the user asks a question (black) the dispatcher chooses one of four routing options
        (\textcolor{tabpurple}{\direct}, \followup, \textcolor{tabred}{\xfquerySkipGT}\ and \textcolor{cyan}{\xfqueryWithGT})
        depending on the question type identified by \questionTrans.
    }
    \label{fig:llm-communication-protocol}
\end{figure*}
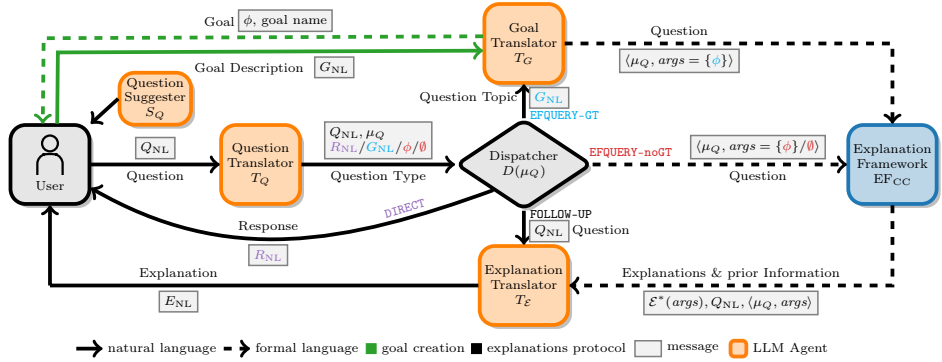

With all components in place, we now describe the implementation of our interactive framework using LLM-based translators to communicate with \explanationframework.
Each translator, as well as the planner and \explanationframework\ (treated as black boxes that generate plans and explanations given a correct input message), is an \emph{agent} that communicates with the others.
We first present the communication protocol between these agents by following the processing of different user question cases, and then discuss the specifics of each translator agent.

As illustrated in Figure \ref{fig:llm-communication-protocol}, everything begins with a question from the user. This question can be selected from the Question Suggester \questionSugg\ or directly expressed in Natural Language. Then, the Question Translator $\questionTrans$ determines whether the question can be answered directly or requires an explanation from $\explanationframework$. To call $\explanationframework$, either $\questionTrans$ or the Goal Translator $\goalTrans$ translates the question from natural to formal language. Then, the formal explanation is translated by the Explanation Translator $\explanationTrans$ and returned to the user.

\subsection{Question Processing} \label{sec:llm-agents-communication-protocol}

Figure~\ref{fig:llm-communication-protocol} shows the communication protocol between the user, translators, and \explanationframework.
From the user's question \questionNL, different routes are possible.
A \emph{Dispatcher} routes the response of $\questionTrans$ depending on the identified question type \questiontype.

\begin{itemize}[left=0pt]
    \item $\questiontype \in \questiontypes$: If $\questionNL$ is classified as one of the question types in $M_Q$. Therefore, an explanation from $\explanationframework$ is necessary, and the resulting formal explanation is then translated by $\explanationTrans$. Nonetheless, depending on the question topic, there are two possibilities:\begin{itemize}[left=0pt]
              \item $\xfquerySkipGT$: The question does not mention goals, or the question argument goals were mapped to known goals (i.e, in $\allgoalsref = \allgoals$). Hence, $\goalTrans$ is bypassed, and the (already known) $\LTLf$\ formula corresponding to the goal is used.
              \item $\xfqueryWithGT$: The goals mentioned in the question cannot be mapped to known goals. Hence, the $\goalTrans$ translates the natural expression of the goal $\goalNL$, creates a new formal goal, and constructs the corresponding $\LTLf$ formula to pass to the $\explanationframework$.
          \end{itemize}

    \item $\questiontype = \direct$: If $\questionTrans$ can answer the question, then the answer is given directly to the user. This includes, for example, questions like ``Which questions can I ask?'' but also incomprehensible messages or unsupported questions. %

    \item $\questiontype = \followup$: If \questionNL\ is identified as a follow-up question, not requiring new input from \explanationframework, then \questionNL\ is directly sent to the \emph{Explanation Translator} $\explanationTrans$.

\end{itemize}

\subsection{LLM Agent Implementations}
\label{sec:translators-implemenation}

All agents are instantiated from the same base model \emph{GPT-4.1-mini}
\footnote{We selected this model as it provided an good balance between inference speed, cost-efficiency, and reasoning performance for real-time interactive tasks.},
with their input and output histories maintained separately. Each LLM agent receives a structured string consisting of the prompts described in the technical appendix and the input components listed in Figure~\ref{fig:llm-communication-protocol}. A translator based only on these inputs can provide only context-independent translations and cannot handle follow-up questions. To enable context-dependent translators, we leverage the LLM context memory, allowing them to build on and reference previous interactions, and thus provide the benefits listed in Section~\ref{sec:iterative_planning_with_explanations}. Although LLM contexts can, in principle, retain all prior user interactions, we maintain separate contexts for each iteration step to avoid mixing information, preserve clarity, and reduce hallucinations. Examples of context-dependent interactions, selections, or summarization of explanations are given in the technical appendix.

Next, we address the tasks of the individual agents and the specific details of their implementation.

\paragraph{Question Suggester}

\questionSugg \ is automatically called to analyze every new iteration step \iterstep.  Its task is to generate one to three relevant questions for the context. These questions are provided to the user via clickable buttons above the input field. To generate the suggestions, it receives the current iteration step context (\goalenf, \goaltrue, \goalfalse, and previous user interactions) and is prompted to propose natural questions that could help. Like all other agents, \questionSugg\ relies on few-shot learning and explicit instructions.

\paragraph{Question Translator}

\questionTrans\ is the first step in question processing. Its tasks go beyond identifying the question type \questiontype. For \direct\, it directly generates the response $\responseNL$. For \xfqueryWithGT\ and \xfquerySkipGT, it extracts one or more goal descriptions \goalNL\ and checks whether they match any $\goalNL' \in \allgoalsref$. Question translations can contain multiple goals, either explicitly ("How can I achieve \textit{S} or \textit{M}?" becomes "$\showqtype \{S,M\}$") or implicitly ("Can I enforce more goals?"  becomes "$\scan \{G_1,G_2,...\}\ \forall G_1,G_2 \in \allgoalsref \setminus \goalenf$"). These are treated by \explanationframework\ as separate requests. \questionTrans\ can detect \followup\ questions to route them to \explanationTrans.

\paragraph{Goal Translator}

The goal translator \goalTrans\ produces both the \LTLf\ formula and a concise natural language description of the goal (used in the UI and in communication between LLM agents). Since our evaluation focuses on the question classification and explanation translation, we use a simple base implementation similar to \emph{End-to-End Approach} of \cite{liu:etal:LRWCoRl22}. The LLM prompt lists the predicates and objects that can compose the literals and includes a few worked examples; see the Appendix for details.

\paragraph{Explanation Translator}

An LLM-based translator can provide user-dependent translation, select explanations, and summarize them. Instead of only providing explanations, we give both conflicts and corrections $\explanations^*(\questionf)$, anticipating follow-up questions (\textit{e.g.} a ``how'' question after a ``why'' question). The translator also receives the user's question, \questionNL, and its formal version, $\questionf$. \questiontype\ indicates how to interpret $\explanations^*(\questionf)$, while \questionNL\ guide generation of a natural response. The translator is expected to use prompt examples to learn how to select or summarize explanations. It is thus a domain-specific design choice to provide examples of the expected selection/summarization strategy. With \followup\ questions, users can explicitly request a summary or selection.

\section{Evaluation}
\label{sec:evaluation}

To evaluate our LLM-based framework, we developed a web-based platform\footnote{Significantly extending \cite{eifler:etal:icaps-22} and an entirely redesigned UI. The source code will be published upon acceptance.}, which enables a human agent to solve iterative planning tasks. This platform is described in details in Appendix~\ref{sec:evaluation-tool}.
We then used this tool to conduct a user study aiming at evaluating the effectiveness of some of our LLM agents and the impact of our LLM-based explanation interface on users' ability to solve a planning task.

Importantly, the goal translator \goalTrans\ was completely disabled during the user study to ensure that all participants were trying to solve the same planning task.

\subsection{User Study Design}
\label{subsec:user-study-design}

Similarly to \cite{eifler:etal:icaps-22}, we used a parent's afternoon planning task as the scenario. The participants were tasked with planning afternoon activities for a family, but due to time constraints, not all of the activities could be carried out. To illustrate, in our running example from Section \ref{sec:running_example}, the parent cannot do both, cook dinner and bring the kids to their match.

As a proxy for user preference, each goal is associated with a utility. The objective of the participants was to select a subset of the goals to maximize the total utility of the plan. In our example, let's say the parent favors their kids, assigning utility $3$ to the match $M$, $2$ to cooking $C$, and $1$ to shopping $S$. Knowing that $M$ and $C$ are in conflict, the
goal subset with maximum utility would be $\{M, S\}$. Naturally, participants' evaluated instance was far more complex.

Participants (recruited via Prolific\footnote{https://www.prolific.com/}) were encouraged to pursue that objective by a bonus reward depending on how close they were to the maximum possible utility. The goal utility is unknown to all other agents (planner, explainer, and LLMs). Consequently, these agents are capable of providing information regarding the solvability of a set of goals, as well as the conflicts and corrections,
but they are unable to provide direct assistance in maximizing utility.

The study was divided into three parts. First, participants were introduced to the tool through a video tutorial and an introduction task ($8$ goals, $2$ conflicts, $6$ corrections). Second, a more complex task ($19$ goals with $224$ conflicts and $313$ corrections) was used to compare the effect of the different interfaces used
on maximizing the utility within $15$ min. Finally, participants completed a post-experiment questionnaire.

To evaluate the effectiveness of LLM agents in an interactive explanation framework, we divided the $131$ participants into two groups. Our control group \templateBased\ included $65$ participants utilizing a \textbf{template-based} interface, which allowed them to select from the predefined set of questions introduced in Section~\ref{sec:goal-onflict-explanation} and receive answers based on simple natural language templates. The second group \LLMBased\ included $66$ participants given the \textbf{LLM-based} interface, which could either select questions from \questionSugg's suggestions, or freely formulate them in natural language.  These questions were processed by LLM-based translators as described in Section~\ref{sec:translator-llm-implementation}.
We did not include a group without explanations, as an earlier study \cite{eifler:etal:icaps-22} had already examined that goal-conflict explanations were more effective than no explanations. More details on the user study design, including the planning instance, can be found in the Appendix~\ref{sec:appendix:user_study}.

\subsection{User Groups Comparison}
\label{subsec:results}

We compared the two groups using quantitative analysis and a feedback questionnaire. Quantitative metrics include the time spent on the task, the maximum utility reached, and the number of questions asked.

\begin{figure}[!t]
    \centering
    \begin{minipage}[t]{0.48\linewidth}
        \centering
        \includegraphics[width=\linewidth]{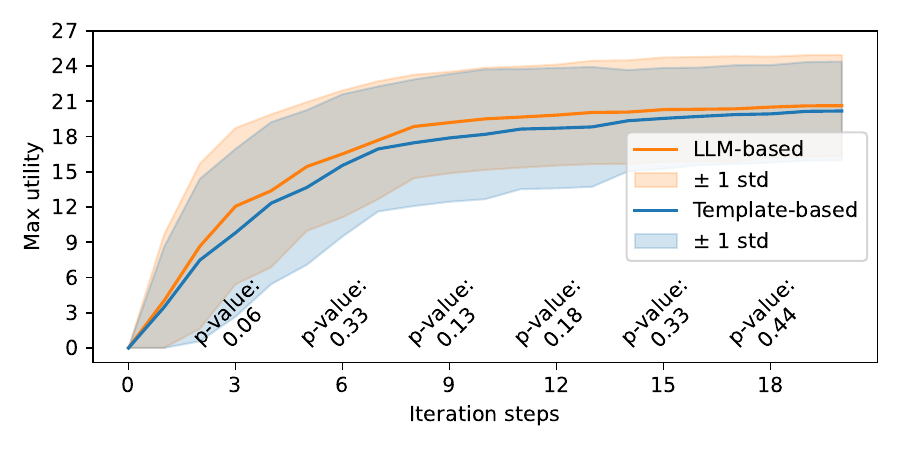}\\[-2ex]
        \caption{
            Comparison of maximum utility achieved over iteration steps between \textcolor{taborange}{\LLMBased}\ and \textcolor{tabblue}{\templateBased}.
            Scores are averaged by group at each iteration step.
        }
        \label{fig:utility_over_iteration_steps}
    \end{minipage}
    \hfill
    \begin{minipage}[t]{0.48\linewidth}
        \centering
        \includegraphics[width=\linewidth]{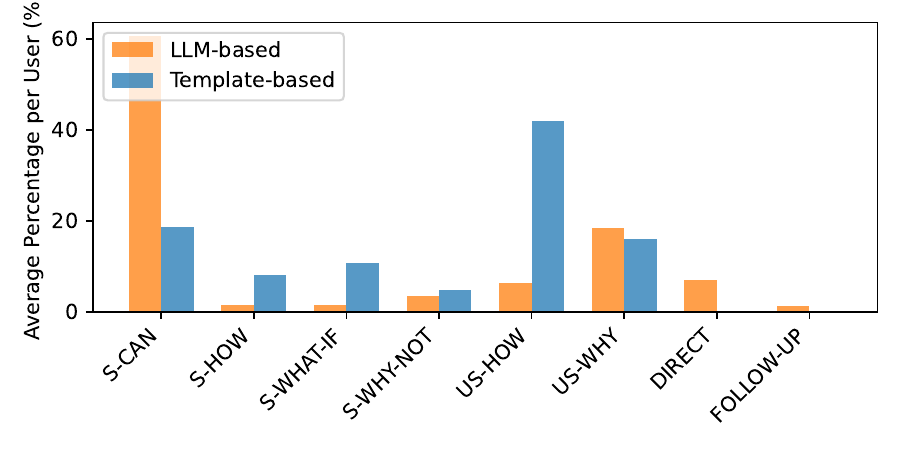}\\[-2ex]
        \caption{
            Comparison of question types used by \textcolor{taborange}{\LLMBased} and \textcolor{tabblue}{\templateBased}. See Sec. \ref{sec:goal-onflict-explanation} for the description of questions.
        }
        \label{fig:question_types}
    \end{minipage}
\end{figure}

\paragraph{Time and maximum utility}
Users took on average $13.9$ mins $\pm$ $2.6$ in \LLMBased\ and $14.0$ mins $\pm$ $2.5$ in \templateBased\ to complete the evaluation task. $10/66$ ($\mathbf{15.2\%}$) users in \LLMBased\ and $9/65$ ($\mathbf{13.8\%}$) in \templateBased\ reached the maximum utility of $27$. On average, \LLMBased\ users achieved a $\mathbf{20.8} \pm 4.3$ utility, and \templateBased\ $\mathbf{20.5} \pm 4.4$. In Figure \ref{fig:utility_over_iteration_steps}, the \LLMBased\ obtain a higher utility with a similar number of steps, in particular in the beginning. However, the variance of scores prevents statistically significant results.

\paragraph{Question types and quantity}
\LLMBased\ asked on average significantly less questions ($\mathbf{11.4}$ $\pm$ $5.9$) vs. ($\mathbf{22.8}$ $\pm$ $19.6$) for \templateBased\, but because the LLM-based question translator can perform several queries to respond a single question, these questions were equivalent to $41.7$ $\pm$ $29.2$ queries to the explanation framework. %
The two groups adopted different question strategies, as shown in Figure~\ref{fig:question_types}. \LLMBased\ users relied a lot more on \scan\ questions than any other type, which may be attributed to the question \textit{Can I enforce any other goals while keeping the current goals enforced?} being often suggested by \questionSugg\ and picked by users; such a question is for \explanationframework\ equivalent to $k$ \scan\ questions if there is $k$ non-enforced goals at the present iteration step.

\paragraph{Question Usage and Achieved Utility Correlation} We conducted a correlation analysis between the number of questions asked and the maximum utility achieved by participants. As shown in Table \ref{tab:correlations}, there is a statistically significant moderate positive correlation for the \LLMBased\ group, indicating that participants who asked more questions tended to achieve higher utility. In contrast, no significant correlation was found for the \templateBased\ group. Note that users who did not use the explanation interface at all are not included in the analysis, so this result does not deny that \templateBased\ explanations are a good baseline. However, it suggests that such explanations do not compound as well as LLM-based conversational explanations. This suggests that users who engaged the most with the LLM-based interface performed better at the task.

\begin{table}[ht]
    \centering
    \setlength{\tabcolsep}{5pt}
    \small
    \begin{tabular}{l c c c c c c c c c}
        \toprule
        \textbf{Group} & \textbf{N} & \multicolumn{2}{c}{\textbf{Pearson: } $U$ vs $Q$} & \multicolumn{2}{c}{\textbf{Pearson: } $U$ vs $Q/it$} & \multicolumn{2}{c}{\textbf{Spearman: } $U$ vs $Q$} & \multicolumn{2}{c}{\textbf{Spearman: } $U$ vs $Q/it$}                                                                   \\
        \cmidrule(lr){3-4} \cmidrule(lr){5-6} \cmidrule(lr){7-8} \cmidrule(lr){9-10}
                       &            & $r$                                               & $p_{\mathrm{perm}}$                                  & $r$                                                & $p_{\mathrm{perm}}$                                   & $\rho$   & $p_{\mathrm{perm}}$ & $\rho$   & $p_{\mathrm{perm}}$ \\
        \midrule
        \LLMBased      & 66         & 0.328                                             & \textbf{0.0068}                                      & 0.254                                              & \textbf{0.0394}                                       & 0.316    & \textbf{0.0106}     & 0.283    & \textbf{0.0238}     \\
        \templateBased & 65         & $-$0.160                                          & 0.1966                                               & $-$0.088                                           & 0.4431                                                & $-$0.004 & 0.9778              & $-$0.016 & 0.9050              \\
        \bottomrule
    \end{tabular}
    \\[0.5em]
    \caption{
        Correlation analysis between achieved utility and question usage.
        $U$ = maximum utility achieved; $Q$ = total questions; $it$ = iteration steps.
        \LLMBased\ shows statistically significant positive associations, indicating that participants asking more questions achieved higher utility.
        \templateBased\ shows no significant associations.
        All $p$-values are permutation-based.
    }
    \label{tab:correlations}
\end{table}
\vspace{-1cm}

\begin{figure}[!t]
    \centering
    \begin{minipage}[c]{0.37\linewidth}
        \caption{
            Feedback questionnaire results comparison for the two groups: \textcolor{tabblue}{\templateBased}\ and \textcolor{taborange}{\LLMBased}. Higher scores indicate stronger agreement. Violin plots show the distribution and mean score of the answers on a Likert scale. \textcolor{taborange}{LLM-based} significantly outperforms \textcolor{tabblue}{template-based} in helping participants to improve their plans and reduce difficulty.
        }
        \label{fig:questionnaire_results}
    \end{minipage}
    \hfill
    \begin{minipage}[c]{0.58\linewidth}
        \centering
        \includegraphics[width=\linewidth]{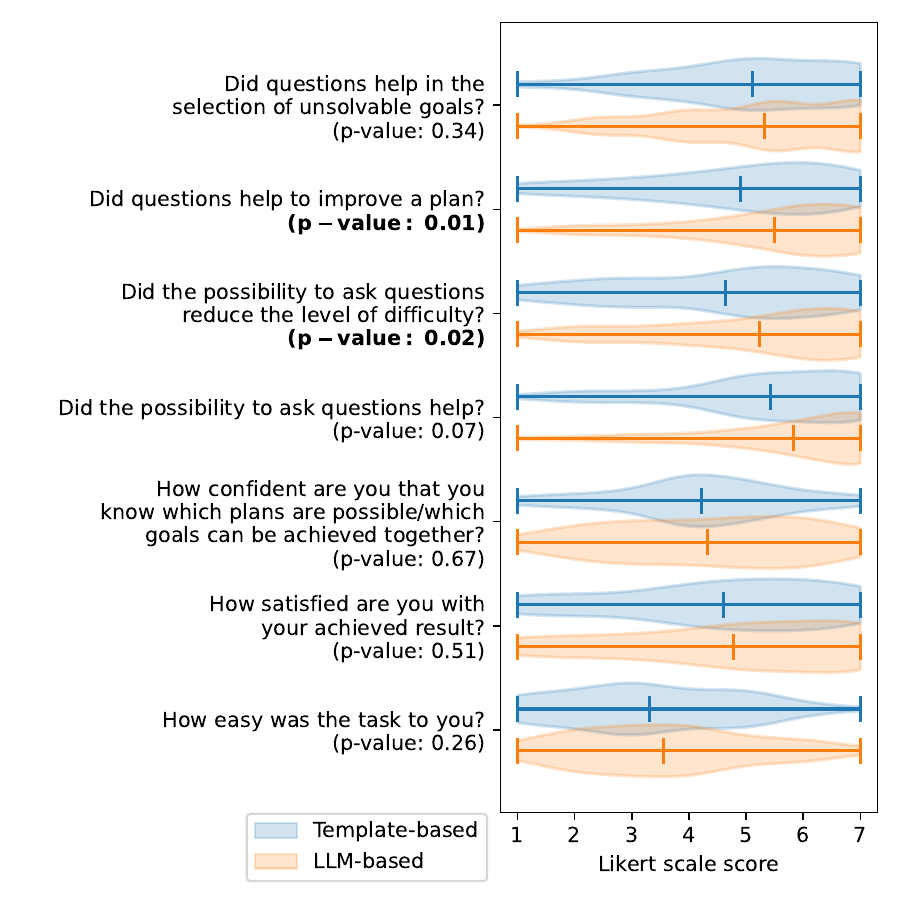}
    \end{minipage}
\end{figure}

\paragraph{Feedback questionnaire}
After completing the evaluation task, participants were asked to fill out a feedback questionnaire assessing the perceived difficulty of the task and the usefulness of the tool. The Figure \ref{fig:questionnaire_results} lists the questions as well as the Likert score \cite{likert1932technique} distributions for the group-question pairs. The \LLMBased\ group mean scores are above those of the \templateBased\ group for all questions. In particular, the two questions on the helpfulness of the explanation interface show statistically significant differences.

\subsection{Evaluation of LLM Agents}

\paragraph{Question Suggester} We first evaluated the usefulness of the question suggester \questionSugg.
The questions picked from suggested questions represented $\mathbf{61.7\%}$ of the 1679 questions collected from \LLMBased, suggesting a strong adoption that did not completely replace manual questions.
\paragraph{Question Translator} We then manually reviewed $100$ randomly selected questions to evaluate the question translator \questionTrans.
The classification of both the question type and the question argument was correct in
$\mathbf{91\%}$ of cases.
Observed failures are mainly bad routing decisions, where the question translator \questionTrans\ unnecessarily asks for clarification or forwards questions directly as a \followup\ question to \explanationTrans\ when a call to \explanationframework\ was required.
These failures typically result in the user asking for an additional clarification.

\paragraph{Other LLM Agents} As described in the introduction of this section, \goalTrans\ was disabled during the user study for practical reasons and so was not evaluated.
As for \explanationTrans\, the main challenge is to measure the correctness of a
summarized explanation, which is out of scope of this paper.

\subsection{Length of Conversations}

\begin{figure}
    \centering
    \includegraphics[width=0.4\linewidth]{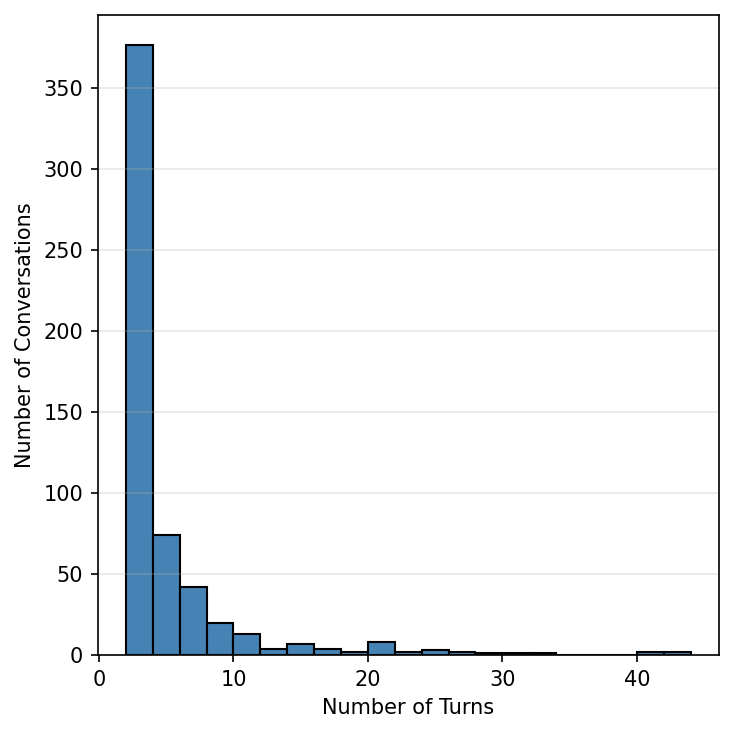}
    \caption{
        Histogram of the number of turns in conversations from \LLMBased\ group.
    }
    \label{fig:turns-distribution}
\end{figure}

Figure~\ref{fig:turns-distribution} shows the distribution of the number of turns in conversations from the \LLMBased\ group. The average number of turns per conversation is \textbf{4.3}, with a standard deviation of \textbf{5.5}. The median number of turns per conversation is \textbf{2.0}, meaning that half of the conversations stopped after just a question and a single response. This also indicates that in half of the iteration steps, users decided that a static explanation was not enough and that engaging in a conversation was required.

\subsection{Dialogue Acts and Explanation Moves Analysis}

Some previous work in natural language processing has focused on the computational analysis of the structure of explanatory dialogues \cite{alshomary-etal-2024-modeling,fichtel-etal-2025-investigating}. We used the models trained by \cite{fichtel-etal-2025-investigating} to annotate conversations obtained from \LLMBased\ with dialogue acts and explanation moves. We report some insights on the structure of conversations in Figure~\ref{fig:dialogue_acts_exp_moves}, as well as additional figures in the appendix. This analysis also shows that while the most common interaction pattern  is a single request for an explanation followed by a single explanation, a significant part of conversations are more complex, with multiple turns and follow-up questions. This suggests that the ability to engage in a conversation and ask follow-up questions is an important feature of the LLM-based explanation interface.

\begin{figure}[htp]
    \centering
    \begin{minipage}[c]{0.49\linewidth}
        \centering
        \includegraphics[width=.99\linewidth]{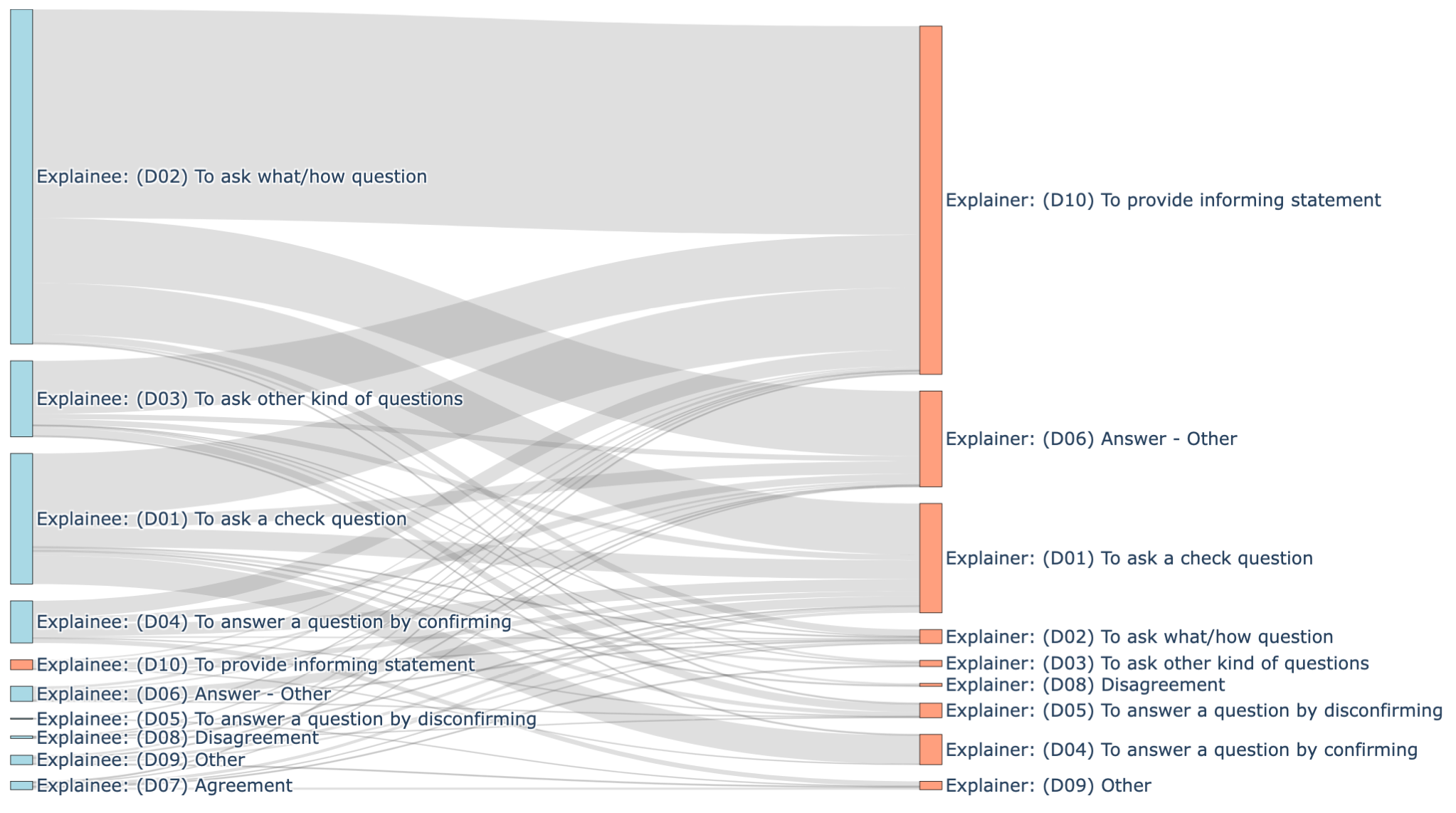}
    \end{minipage}
    \hfill
    \begin{minipage}[c]{0.49\linewidth}
        \centering
        \includegraphics[width=.99\linewidth]{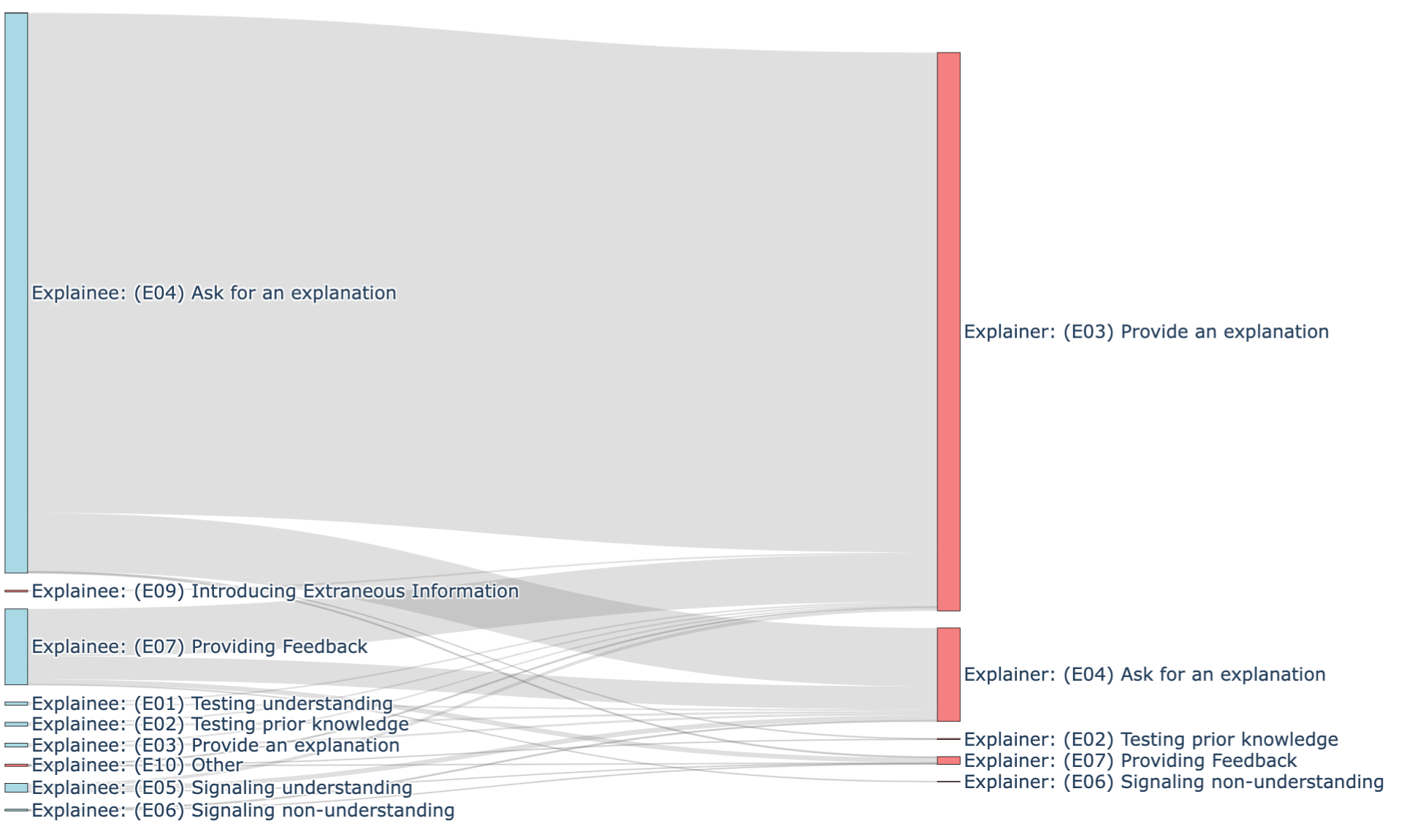}
    \end{minipage}
    \caption{
        Dialogue acts (left) and explanation moves (right) distribution in \LLMBased\ conversations using model from \cite{fichtel-etal-2025-investigating}. The flow indicates which dialogue acts (respectively explanation move) usually follow user's dialogue act (respectively explanation move).
    }
    \label{fig:dialogue_acts_exp_moves}
\end{figure}

\subsection{Limitations and Discussion}
\paragraph{Instance and Task} While our user study provides evidence of better performance of \LLMBased\
over \templateBased, we note that these results were obtained on a
single planning task instance of moderate difficulty.
Although we chose this setup to align with a previous user study
in which template-based goal-conflict explanations were found useful,
it remains an open question how the proposed architecture would help
on a wider range of problems.  Another limitation of our study is that it was conducted with lay users.
Whether similar results hold for the target users of such tools (domain experts)
remains open.
We suspect that domain experts will ask more complex questions than lay persons, leveraging more complex capabilities of the LLM-based interface. The positive correlation between question usage and achieved utility in the \LLMBased\ group supports this hypothesis.

\paragraph{Explanation Interface is also about UI}
We used two explanation interfaces to compare high-level features: a chat-based interface against a more traditional GUI. However, more practical features, like the design of menus and buttons, the presence of visual elements, or other UX-related elements, also could have a significant impact. While we did not focus on this aspect, we used a pilot study to help design a good UI for both interfaces; the pilot study led us to include the question suggester, for example.
Notably, many participants from both groups voluntarily commented after the study that they found the interaction enjoyable.

\section{Conclusion and Future Work}
\label{sec:conclusion}

We have found evidence of a significant subjective improvement when using LLM-based explanation interfaces compared to the baseline interface. Moreover, we observed a consistent trend suggesting that this subjective gain may be supported by an objective improvement, namely a convergence toward an acceptable plan in fewer iteration steps. We also observed a significant positive correlation between the adoption of our interface and performance on the planning task. These results suggest potential for LLM-based interfaces, particularly for users who actively engage with them.

Future work will assess LLM-translated explanations on a wider range of tasks, including more realistic planning uses cases with end users.
In such contexts, the goal translator will becomes a more important part of our framework and will be evaluated accordingly. Similarly, we will carefully assess summarization capabilities of LLMs and their capacity to respond to different users requirements.

We also plan on investigating the extension of this approach to other explanations
frameworks.
This includes MUS and
MCS based approaches used in Constraint Programming
\cite{poveda:etal:pthg24,gamba:etal:jar23} but also other approaches used
in planning \cite{krarup:etal:jair21,sreedharan:etal:ai21}.
In addition to the classification of the question type, this adds the challenge of deciding
which explanation approach is most suitable to answer the question.

\begin{credits}

  \subsubsection{\ackname}  This work was funded by the European Union’s Horizon Europe Research and Innovation program under the grant agreement TUPLES No 101070149, and was supported by the Artificial and Natural Intelligence Toulouse Institute (ANITI). ANITI is funded by the France 2030 program under the Grant agreement ANR-23-IACL-0002.

  \subsubsection{\discintname}
  The authors have no competing interests to declare that are
  relevant to the content of this article.
\end{credits}
\bibliographystyle{splncs04}
\bibliography{biblio.bib}

\clearpage
\appendix

\section{Platform for Iterative Planning with Explanations}
\label{sec:evaluation-tool}

\noindent
\emph{The source code is available at https://github.com/r-eifler/IPEXCO-frontend.}\\
\noindent
First we describe the platform architecture, then we give an overview over the
supported features covering the iterative planning process, the interface options
and the adaptations for conducting user studies.

\subsection{Architecture}

The tool is a web platform, meaning it runs directly in the browser and does not
require installation on individual users' machines.
This feature is especially beneficial for conducting user studies.

\begin{figure}[!htp]
    \centering
    \includegraphics[scale=0.6]{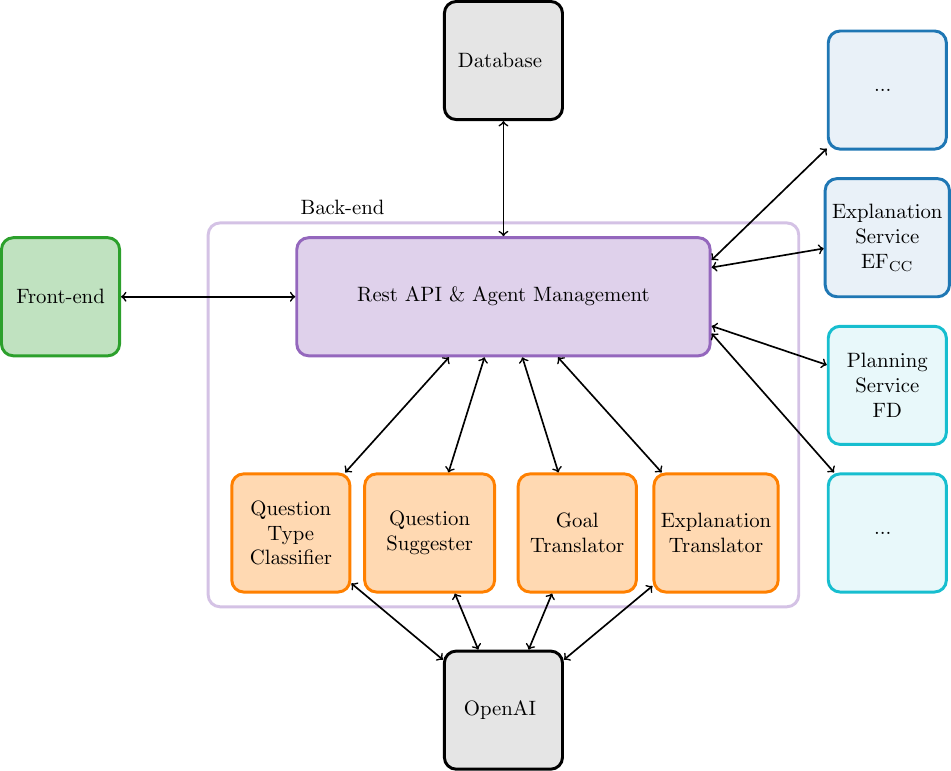}
    \caption{Architecture of the web platform}
    \label{fig:ipexco-architecture}
\end{figure}

We are using Angular\footnote{https://angular.dev/} for the
front-end and Node.js\footnote{https://nodejs.org} with express\footnote{https://expressjs.com/}
and a MongoDB\footnote{https://www.mongodb.com/} database for the back-end.
Figure~\ref{fig:ipexco-architecture} depicts the architecture of the platform.

The system has a modular architecture.
Explanations and plans are  provided by individual services that in the future
can be extended with explanations and plans based on different approaches.
The back-end and the explanation and planning services implement REST APIs.
The communication between the back-end and a service is asynchronous.
This means the service notifies the back-end when a job has been completed.
The services themselves maintain a queue for incoming jobs and schedule
them based on resource availability.

The currently used planner service is based on Fast Downward \cite{helmert:jair-06}
extended with a compilation approach to handle temporal goal as described in \cite{eifler:etal:ijcai-20}.
The explanations service computes the MUS and the MCS using the algorithms introduced
in \cite{eifler:etal:ijcai-20} implemented on top of Fast Downward \cite{helmert:jair-06}.

\subsection{Planning Process}

The main feature of the platform is the iterative planning process with explanations.
In the following we explain how a planning task is provided, how goals can be
defined and how the interface for the iterative planning process and the
explanations is designed.

\subsubsection{Planning Project}

Each planning process starts with the creation of a project.
The planning task is given by a PDDL domain and problem file.
The domain definition, the actions, and the initial state are fixed, they
cannot be modified during the planning process.
From the problem file only the initial state definition is considered,
the goals are disregarded.
The exploration of the plan space is guided by the enforced goals.
Goal are defined during the planning process via different interface
options, which are described in the following section.

In general the platform focuses on domain independent features.
However, for some features (\eg\ templates for goals and prompts for the LLM-based
translators) domain dependent information can be added.
Thus, each project is associated with a domain, \eg\ \texttt{blocksworld},
\texttt{transport} or \texttt{parents-afternoon}.

\subsubsection{Goal Creation}

At the beginning of the planning process the user must define all known goals
and potential preferences they are already aware of.
During the planning process, after having inspected some samples plans,
the goals can be extended to account for any new goals and preferences.

Goals are defined by \LTLf\ formulas with literal based on the predicates and
objects of the planning task.
In order to facilitate the creation of goals, the requirement to write \LTLf\
is circumvented, as it is not suitable for laypeople and is generally an error-prone
process.
Two options are supported.

\begin{figure}[!th]
    \centering
    \begin{tikzpicture}
        \node[] (tmp) at (0,0) {
            \includegraphics[width=0.35\linewidth]{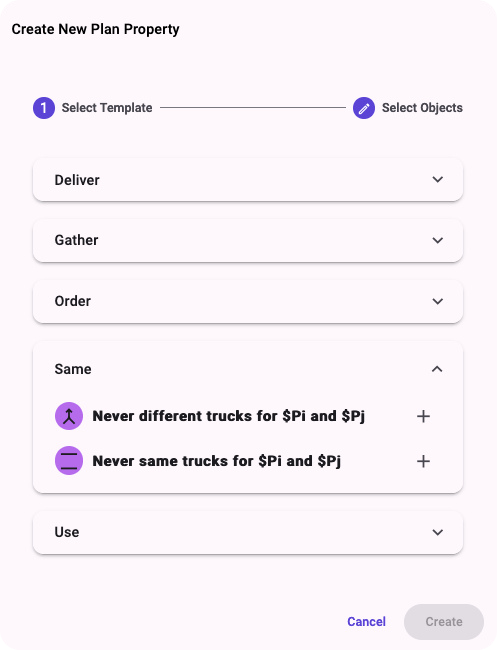}
        };
        \node[] (tmp) at (9.2,0.25) {
            \includegraphics[width=0.35\linewidth]{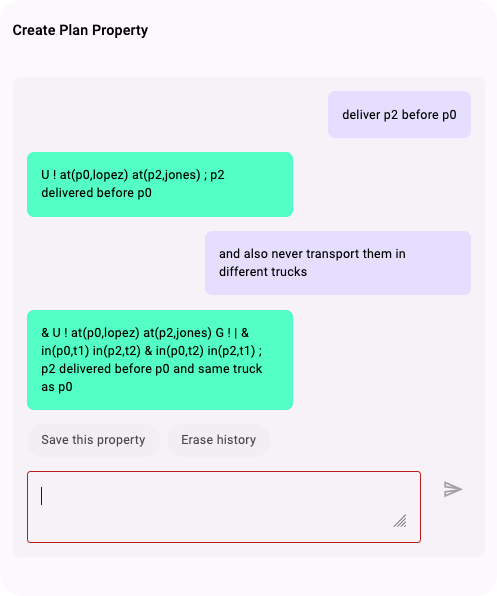}
        };
    \end{tikzpicture}
    \caption{Screenshot of the goal creation using templates baseline (left) and LLM-based goal translator (right).}
    \label{fig:tool-goal-creation}
\end{figure}

\paragraph{Templates}
It is possible to define templates for each domain to cover commonly used
goals and temporal preferences.
Such a template maps a natural language description, \eg\
``Load package \texttt{\$P}$_\texttt{i}$ before package \texttt{\$P}$_\texttt{j}$
into truck \texttt{\$T}'' to an \LTLf\ formula $\LTLnot in(P_j,T) \LTLuntil in(P_i,T)$.
To instantiate a goal on the basis of a template, it is then only necessary to
select the specific objects, in this example the two packages and the truck.
To restrict the selection of objects offered for selection the allowed types
(e.g. \texttt{\$P}$_\texttt{i}$ must be of type \texttt{package})
and facts that must/ must not be satisfied in the initial state
(e.g. $\LTLnot in(P_j,T)$) can be specified.
This helps to ensure that only well-formed and meaningful goals can be instantiated.

In Figure~\ref{fig:tool-goal-creation} on the left the list of available templates
for a transport domain is show.
The first step in creating a goal is choosing one of those templates.
In the second step the objects are selected.

\paragraph{LLM-Based Goal Translator}
The second option is to use a LLM-based goal translator, that translates a
natural language description of the goal into an \LTLf\ formula.
An example is given on the right in Figure~\ref{fig:tool-goal-creation}.
The LLM infers the delivery location from the planning task.
The multi-step interaction to refine the goal is only possible with a context-dependent translator.
The translated \LTLf\ formulas shown in this figure are intended to be shown
exclusively to expert users.
An alternative approach involves the use of a \textit{reverse translation} of
the users' question, which is then converted back into natural language without ambiguities.
(e.g.``deliver p2 before p0'' $\rightarrow$ \textit{I understood this goal as "p0 is not delivered to lopez until p2 is delivered to jones"}).
The summary of the goal description, here ``p2 delivered before p0 and same
truck for p0'' is used as the description of the goal in the interface.

\paragraph{Goal Properties and Visualization}

\begin{figure}[!th]
    \centering
    \includegraphics[scale=0.5]{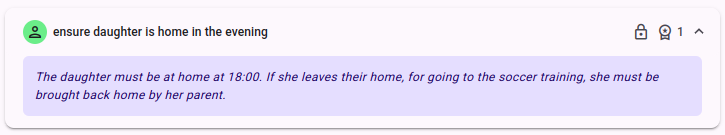}
    \caption{Screenshot of an open goal panel.}
    \label{fig:goal-panel}
\end{figure}

In the interface goals are depicted as shown in Figure~\ref{fig:goal-panel}.
Each goal has a natural language description and is associated with a color and icon.
These can be used to visually group goals by for example the type of temporal
property (\eg\ delivery order) or the involved objects (\eg\ truck $T_1$).
If the panel is opened it gives access to a more detailed description of the
goal.

Goals can be \emph{locked} (indicated by a small lock as shown in the top right in Figure~\ref{fig:goal-panel}).
This means they are automatically enforced in each iteration step.

\subsubsection{Iterative Planning Process}

The iterative planning process interface must support the user in taking an
informed decision about which goals they should enforce in the next iteration.
To take this decision, users need an overview of the existing iteration steps and
access to more detailed information for each step.
At the same time user should be able to select the enforced goals for the next
iteration step.
Our tool implements this process as follows.

\begin{figure}[!th]
    \centering
    \includegraphics[width=0.8\linewidth]{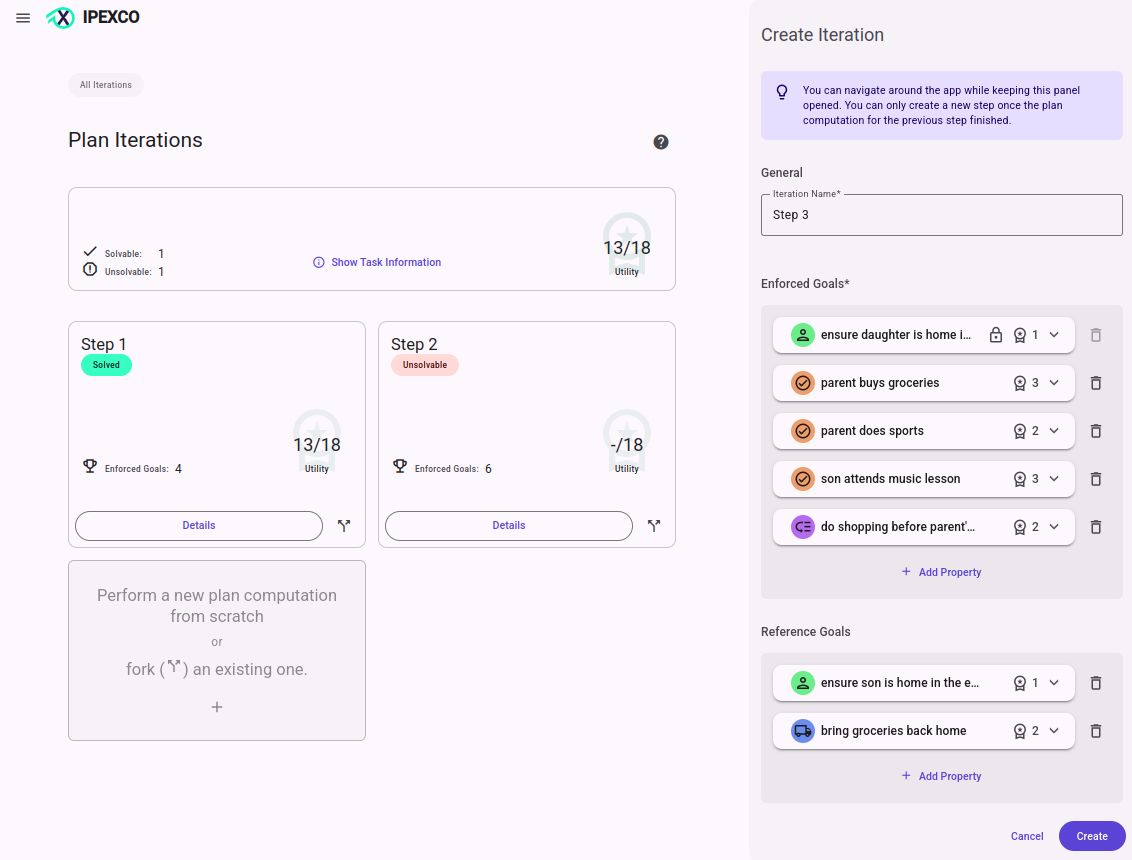}
    \caption{Screenshot of the iteration steps overview left with the interface
        to create new steps on the right.
    }
    \label{fig:tool-iteration-steps}
\end{figure}

The overview of the iteration steps is shown in the left in Figure~\ref{fig:tool-iteration-steps}.
It allows to quickly identify which steps were solvable and unsolvable
and the utility they achieved (More about the utility in Section~\ref{sec:tool-user-study-adaptations}).

\begin{figure}[!th]
    \centering
    \begin{tikzpicture}
        \node[] (tmp) at (0,0) {
            \includegraphics[width=0.4\linewidth]{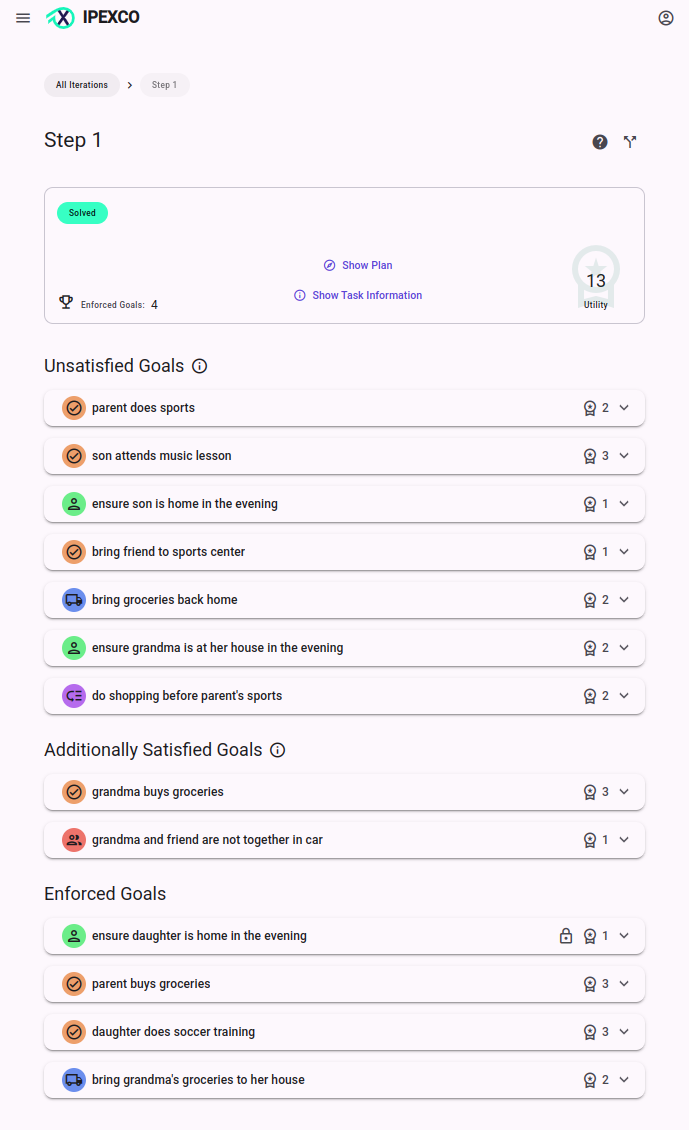}
        };
        \node[] (tmp) at (9,0) {
            \includegraphics[width=0.4\linewidth]{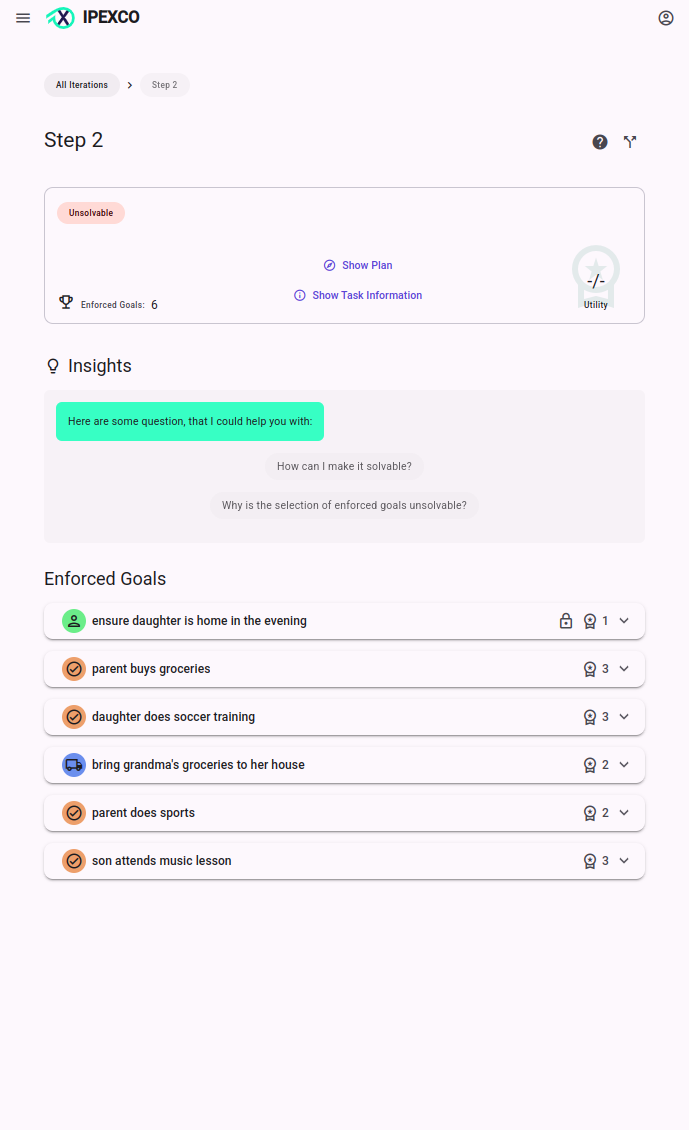}
        };
    \end{tikzpicture}
    \caption{Screenshot of the details view of a solvable iteration step (left)
        and an unsolvable step (right).}
    \label{fig:tool-iteration-steps-details}
\end{figure}

The details view for a step is shown in Figure~\ref{fig:tool-iteration-steps-details}.
In addition to the information in the overview it gives access to the
following information:

\begin{itemize}
    \item{Solvable Step $\iterstep_i$}:
          \begin{itemize}
              \item the plan: $\plan_i$
              \item the additional satisfied reference goals: $\goaltrue(\plan) \setminus \goalenf_i$
              \item the not satisfied reference goals:  $\goalfalse(\plan)$
              \item the enforced goals: $\goalenf_i$
          \end{itemize}
    \item{Unsolvable Step $\iterstep_i$}:
          \begin{itemize}
              \item the enforced goals: $\goalenf_i$
          \end{itemize}
\end{itemize}

Users can create the next iteration step by either starting from scratch or by
forking from an existing step.
To allow the creation of the next step while simultaneously having access to
the information described above and the explanations (presented in the next Section),
the interface is located in a sidebar as shown on the right in Figure~\ref{fig:tool-iteration-steps}.

To create a new step a user must select at least one enforced goal.
This can be an already existing or a newly created goal.
In addition, the user can also select or create new reference goals $\goalref_{i+1}$.
If the user is satisfied with their selection they can create a new step.
This automatically triggers the computation of a plan that must satisfy
the enforced goals.

\subsubsection{Explanations Interfaces}

Explanations are provided in the details view of the iteration steps.
The computation of an explanation is triggered as soon as a question is
asked.
If a question can be answered based on the conflicts or corrections from
a previous question, then the cashed result is used instead.
The access to the explanations and their representation differs depending on the
chosen interface.
Currently,  two options are supported.

\paragraph{Template Based}

\begin{figure}[!th]
    \centering
    \begin{tikzpicture}
        \node[] (tmp) at (0,0) {
            \includegraphics[width=0.45\linewidth]{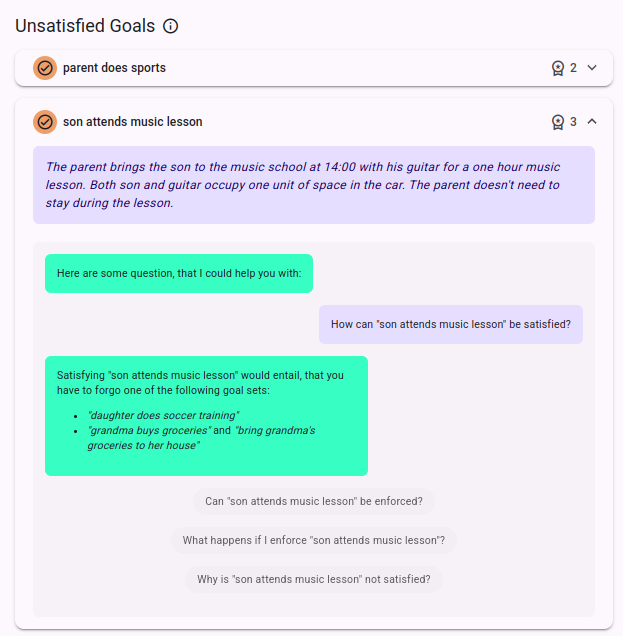}
        };
        \node[] (tmp) at (9.2,0) {
            \includegraphics[width=0.45\linewidth]{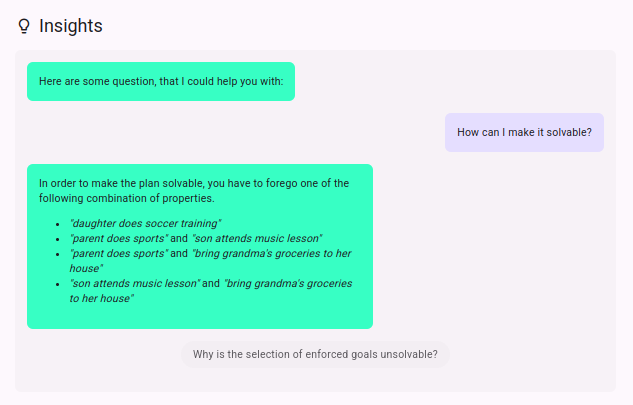}
        };
    \end{tikzpicture}
    \caption{Screenshot of the template-based explanations for a solvable step (left) and
        an unsolvable step (right).}
    \label{fig:explanations-template-based}
\end{figure}

In the template based explanation interface the user is restricted to a predefined
list of questions and the answer simply lists all explanations.
To allow for a similar experience as in the LLM-based version described next,
the questions and explanations are depicted in a chat interface as shown in
Figure~\ref{fig:explanations-template-based}.

For an unsolvable step the questions refer to the entire selection of
enforced goals.
Thus, the explanation interface is displayed above the list of enforced
goals (see right Figure~\ref{fig:tool-iteration-steps-details} and Figure~\ref{fig:explanations-template-based}).

In a solvable step the questions refer to individual unsatisfied reference
gals $\goalfalse(\plan)$.
These goals are the arguments of the questions.
Thus, for each unsatisfied reference goal there is a separate explanation
interface accessible by clicking on the goal panel
(see left Figure~\ref{fig:explanations-template-based}).

\paragraph{LLM-Chat Based}

\begin{figure}[!th]
    \centering
    \begin{tikzpicture}
        \node[] (tmp) at (0,0) {
            \includegraphics[width=0.90\linewidth]{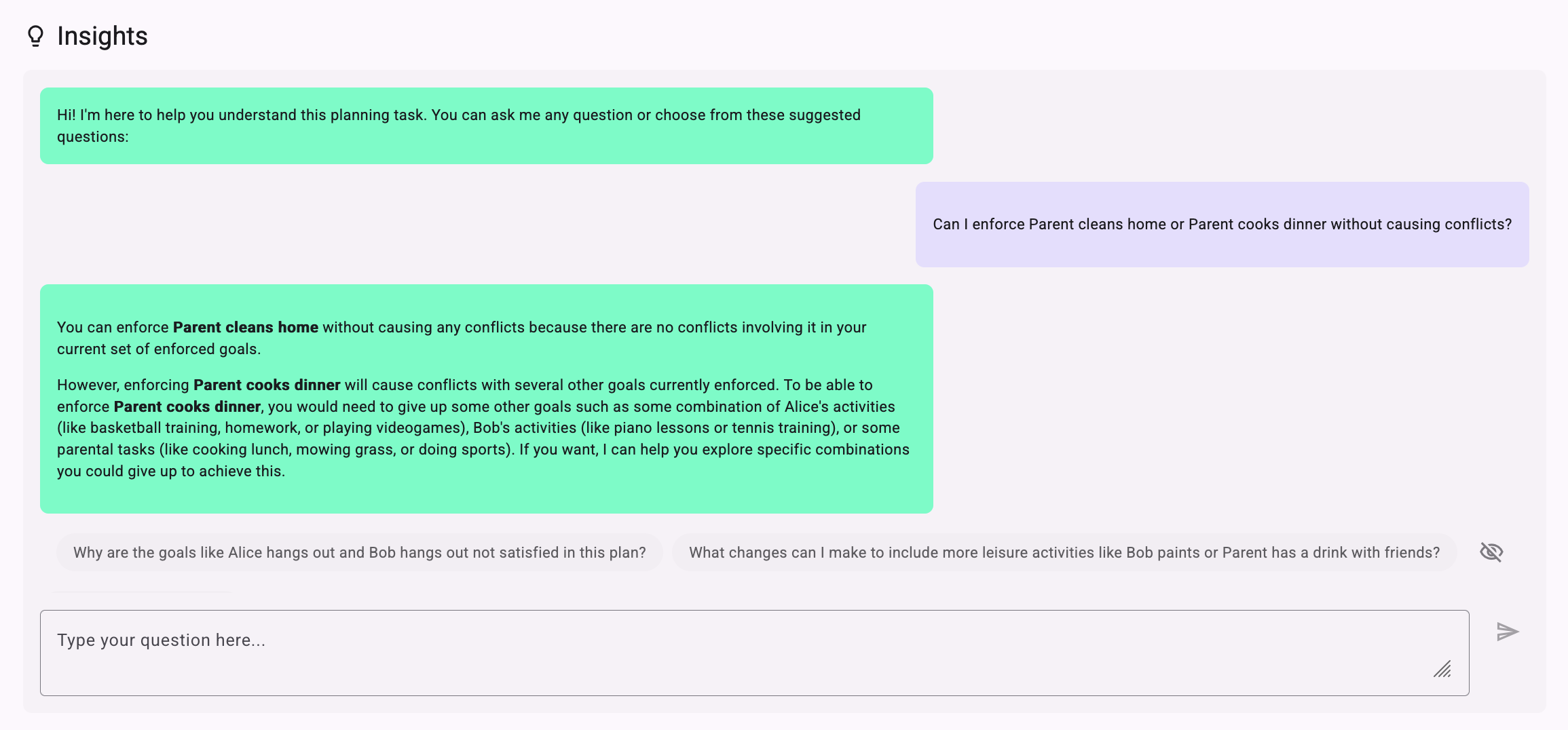}
        };
    \end{tikzpicture}
    \caption{Screenshot of the LLM-based explanations.}
    \label{fig:explanations-llm-based}
\end{figure}

The LLM-Chat based explanation interface is displayed in the top of the
iteration step details view for both solvable and unsolvable steps
(see Figure\ref{fig:explanations-llm-based}).
More examples can be found in Section~\ref{sec:appendix:llm-examples}.
Users can freely formulate questions.
If the question is answered by calling \explanationframework\ first
the user is notified how the question was understood.
This is implemented by the question translator to \textit{reverse translate} the
question similarly to the LLM-based goal translator described earlier.
Then the response from the explanation translator is displayed.
In case of direct answers by the question translator or answers to
follow-up questions from the explanation translator, the answer is given
directly.

\FloatBarrier
\subsection{User Study Adaptations}
\label{sec:tool-user-study}

In order to facilitate the execution of a user study within a controlled
environment, the following adaptations were made.

\paragraph{Demo Environment}

The computation of explanations requires a significant amount of time.
In order to provide a responsive experience for the users, all conflicts and
corrections are precomputed.
This requires to fix the set of reference goals.

A \emph{demo} is created based on a project by selecting a set of goals.
Each goal is assigned a utility that acts as a proxy for an intrinsic
preference.
Then all MUS and MCS for these goals are computed.

Fixing the set of reference goals also provides a controlled environment for
a user study.
All users must solve the same optimization problem, which allows to
use the maximal achieved utility as a metric.

\paragraph{Adaptations Iterative Planning}
\label{sec:tool-user-study-adaptations}

During a user study some features of the iterative planning process are disabled.
Users do not have access to the sample plans, but only to the information
displayed in the iteration step detail view.
Since the goals are fixed, users cannot create any new goals.
If a user creates a new iteration step, then they must only select the enforced
goals.
All not enforced goals are automatically selected as reference goals.
To motivate users to perform well, to achieve a high utility, the maximal
achievable utility for the demo and the utility of the best iteration step are displayed.

\begin{figure}[!th]
    \centering
    \includegraphics[width=0.8\linewidth]{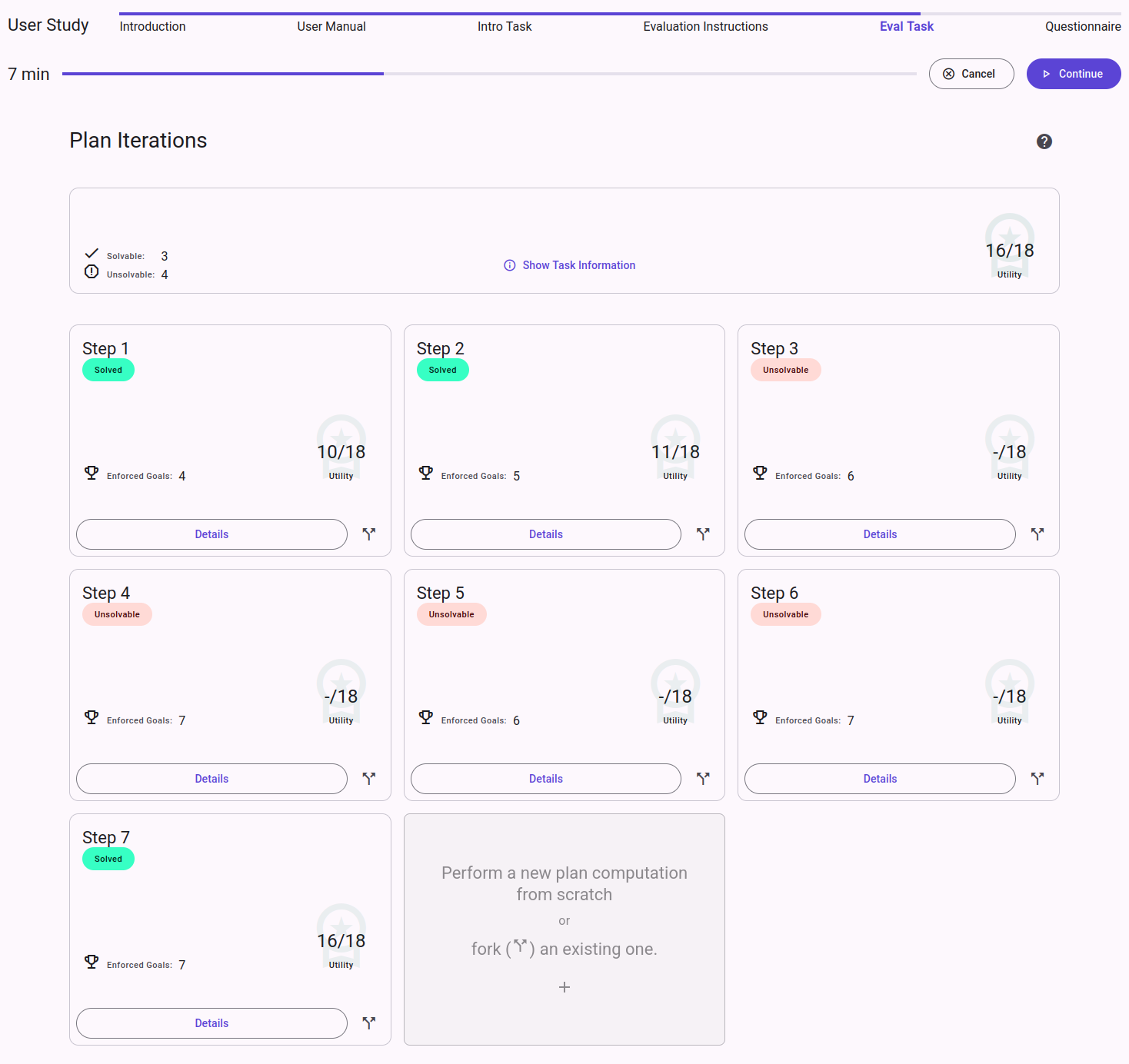}
    \caption{Iteration steps list for a demo during the user study.}
    \label{fig:iteration-steps-demo}
\end{figure}

Figure~\ref{fig:iteration-steps-demo} shows an example list of iteration
steps for the demo used in the user study.
Overall a utility of $16$ out of $18$ is achieved.
In the top the individual parts of the user study are shown and below the
timer for the available processing time.

\FloatBarrier

\section{User Study}
\label{sec:appendix:user_study}

\subsection{Detailed Process}
\label{subsec:appendix:user_study_process}

We recruited (after filtering) $131$ fluent in english participants through Prolific (\url{www.prolific.com}), keeping only the submissions of participants that reached the end of the user study, achieved at least two iteration steps including a solvable one, and used the explanation interface at least one.
The estimated time to complete the study was between $30$ and $35$ minutes.
Participants were randomly assigned to one of two groups,
resulting in $65$ participants in \templateBased\ and $66$ in \LLMBased.

The participants were rewarded with a base payment of \textsterling $4.20$ \ and a bonus payment of \textsterling $0 $ \ to \textsterling $5$ depending of achieved maximum utility.

The user study was divided into three main parts:

\paragraph{Introduction to the Task}
After consenting to participate in our online user-study,
the tool and its possibilities were presented via a 1mn30 video tutorial.
Afterwards, participants were presented a toy example of the Parents Afternoon domain,
more details in Section~\ref{subsec:appendix:user_study_instance}.
Subsequently, they were given 10 minutes to interact with the system and
explore the explanation interface.
The objective of this first part was to familiarize participants with the tool and the domain.
If they reached the maximum utility before 10 minutes,
they were proposed to continue with the next part.

\paragraph{Evaluation Scenario}
The second part consist of a more complex scenario,
as described in Section~\ref{subsec:appendix:user_study_instance}.
This scenario was introduced to the participants similarly to the toy example.
Then they were given 15 minutes to interact with the system.
Users were aware of the maximum achievable utility and notified when they reached it.

\paragraph{Post-experiment Questionnaire}
Finally, participants were asked to fill out a questionnaire to evaluate.
Question details and summary of answers are presented in Section~\ref{subsec:appendix:user_study_results}.
The evaluated points were:
\begin{itemize}
    \item The perception of the task difficulty;
    \item The perception of the usefulness of the question interface;
\end{itemize}

\subsection{Instance}
\label{subsec:appendix:user_study_instance}

\paragraph{Introduction instance specifics} :
\begin{itemize}
    \item People: a parent (the only one who can drive the car), alice and bob (kids)
    \item Items: groceries (required for cooking lunch and dinner) tennis racket (required for tennis training)
    \item Locations: home, groceries store, sport center, and city center
    \item Activities (utility) :
          do groceries  (1) ,
          clean home (1),
          cook lunch (4),
          cook dinner (3),
          bob does tennis training (2),
          bob does homework (3),
          alice does violin lesson (1),
          alice does homework (2),
    \item Time: tracked in 10 levels (08h00-18h00)
    \item The car has a capacity of 2, and initially everyone is free (not busy) at home.
\end{itemize}

There is only two global conflicts ; removing the goal with the least utility in these two conflicts (alice's violin lesson and clean home) was the optimal solution.

The maximum utility was 15.

\paragraph{Evaluation instance specifics}:
\begin{itemize}
    \item \textbf{People:} a parent (the only one who can drive the car), Alice and Bob (kids)

    \item \textbf{Items:} none currently active

    \item \textbf{Locations:} home, gymnasium, city center, and music center

    \item \textbf{Activities (with utility):}
          \begin{itemize}
              \item \textit{Alice's activities:}
                    \begin{itemize}
                        \item does homework [3] (2h, 08:00--10:00, at home)
                        \item hangs out [2] (2h, 11:00--13:00, at city center)
                        \item plays video games [2] (2h, 11:00--13:00, at home)
                        \item does basketball training [3] (2h, 14:00--16:00, at gymnasium)
                        \item takes violin lesson [2] (1h, 17:00--18:00, at music center)
                    \end{itemize}
              \item \textit{Bob's activities:}
                    \begin{itemize}
                        \item does homework [3] (2h, 08:00--10:00, at home)
                        \item hangs out [2] (2h, 11:00--13:00, at city center)
                        \item paints [2] (2h, 11:00--13:00, at home)
                        \item does tennis training [2] (2h, 14:00--16:00, at gymnasium)
                        \item takes piano lesson [3] (1h, 17:00--18:00, at music center)
                    \end{itemize}
              \item \textit{Parent's activities:}
                    \begin{itemize}
                        \item cleans home [2] (2h, 08:00--10:00, at home)
                        \item does laundry [1] (1h, 10:00--11:00, at home)
                        \item mows grass [1] (1h, 11:00--12:00, at home)
                        \item cooks lunch [1] (1h, 12:00--13:00, at home)
                        \item cooks dinner [1] (1h, 17:00--18:00, at home)
                        \item does sports [2] (2h, 14:00--16:00, at gymnasium)
                        \item goes to the hairdresser [1] (1h, 14:00--15:00, at city center)
                        \item has a drink with friends [3] (2h, 15:00--17:00, at city center)
                        \item goes shopping [1] (1h, 17:00--18:00, at city center)
                    \end{itemize}
          \end{itemize}

    \item \textbf{Time:} tracked in 11 levels (08:00--18:00, i.e., 10 hours)

    \item \textbf{Constraints:} The car has a capacity of 3, and initially everyone is free (not busy) at home.
\end{itemize}

There was 224 Conflicts and 313 Corrections.

The maximum utility was 27 and was obtained by removing : Alice hangs out, Bob hangs out, Parent cooks dinner, Parent goes shopping, Parent goes to the hairdresser, Parent has a drink with friends.

\subsection{Additional Results}
\label{subsec:appendix:user_study_results}

\subsubsection{Utility over Time}

\begin{figure}[!ht]
    \centering
    \includegraphics[width=0.8\linewidth]{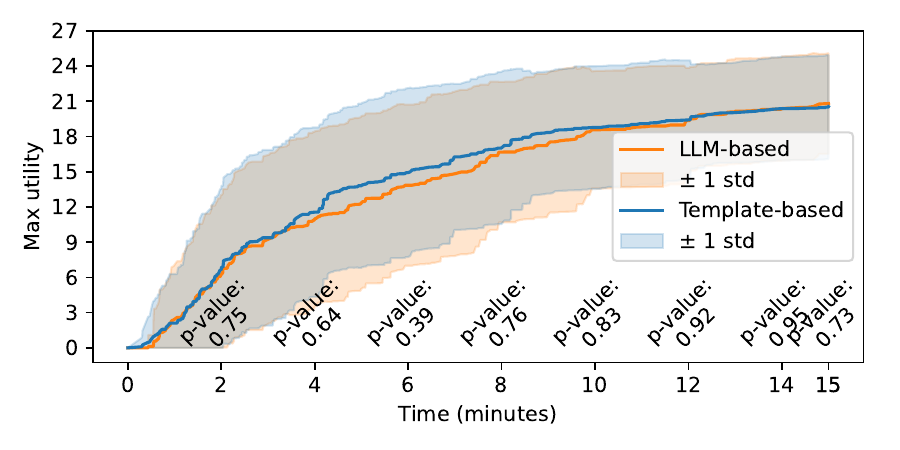}
    \caption{
        Comparison of max utility achieved over time between
        the two groups: \templateBased\ and \LLMBased.
    }
    \label{fig:utility_over_iteration_steps_appendix}
\end{figure}

Similarly to the comparison of the utility score over iteration steps,
we also analyzed it across time. However, no significant difference was observed (see Figure~\ref{fig:utility_over_iteration_steps_appendix}). We can hypothesize that the time benefits of achieving a higher utility per iteration steps were at least partially balanced by users spending more time in each iteration step engaging in a conversation with the explanation interface. This does not negate the benefit of the LLM-based interface since reducing the number of iteration steps can be beneficial in real-world scenarios where each iteration step may require a costly re-planning process. Also, spending more time in an iteration step may indicate that users are building a better understanding of the problem and the solution space, which can be beneficial for their overall experience and trust in the system.

\subsubsection{Non-Averaged Utility over Steps}

\begin{figure}[htbp]
    \centering
    \begin{subfigure}[b]{0.48\textwidth}
        \includegraphics[width=\textwidth]{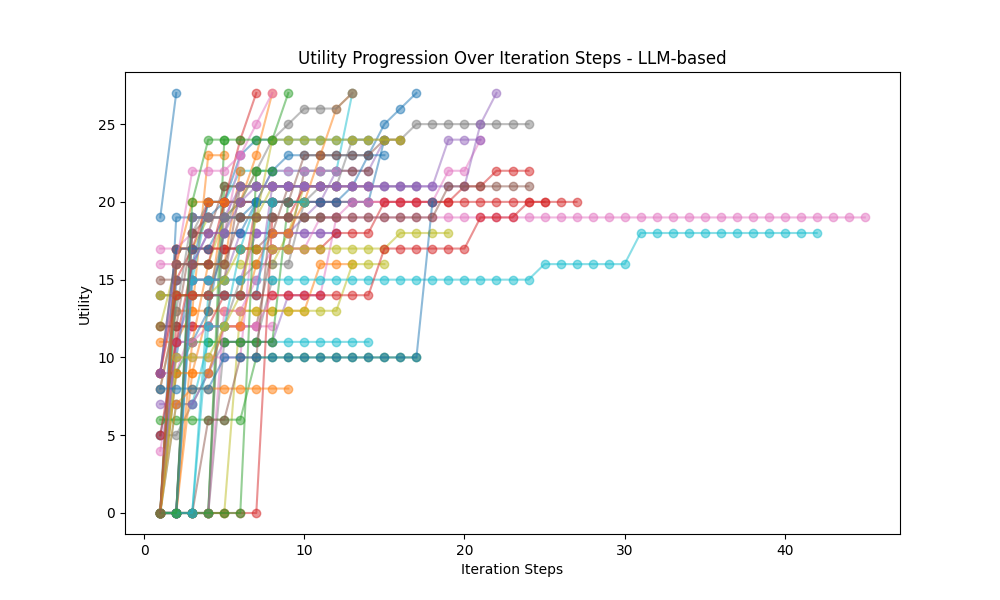}
        \caption{LLM-based interface}
    \end{subfigure}
    \hfill
    \begin{subfigure}[b]{0.48\textwidth}
        \includegraphics[width=\textwidth]{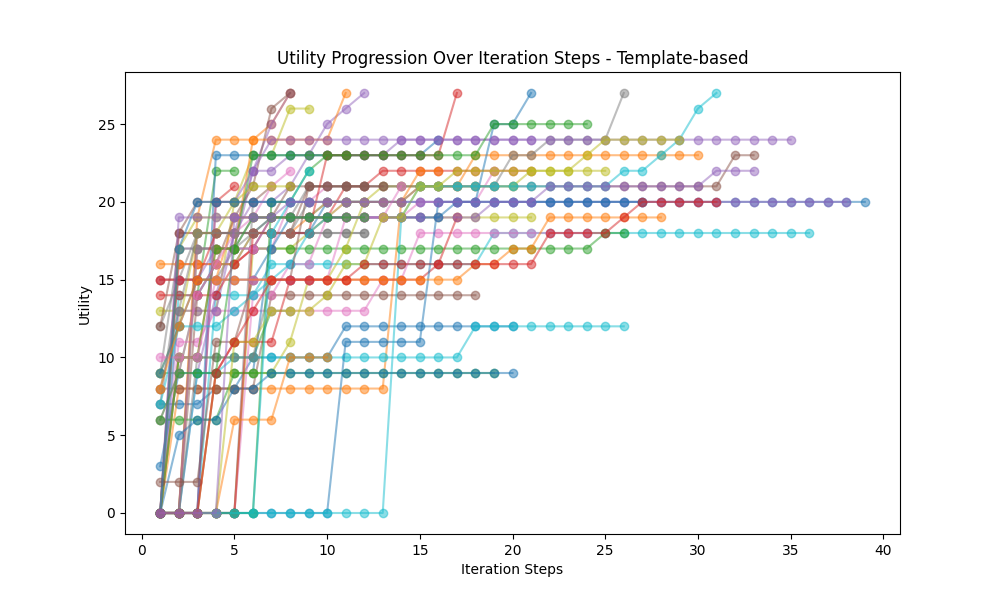}
        \caption{Template-based interface}
    \end{subfigure}
    \caption{}
    \label{fig:max-achieved-unaveraged}
\end{figure}

We also provide (Figure~\ref{fig:max-achieved-unaveraged}) non-averaged figures of the maximum achieved utility over iteration steps: each accepted participant of the user study is one trajectory.

\subsubsection{Dialogue Acts and Explanation Moves Analysis}

\begin{figure}
    \centering
    \includegraphics[width=\linewidth]{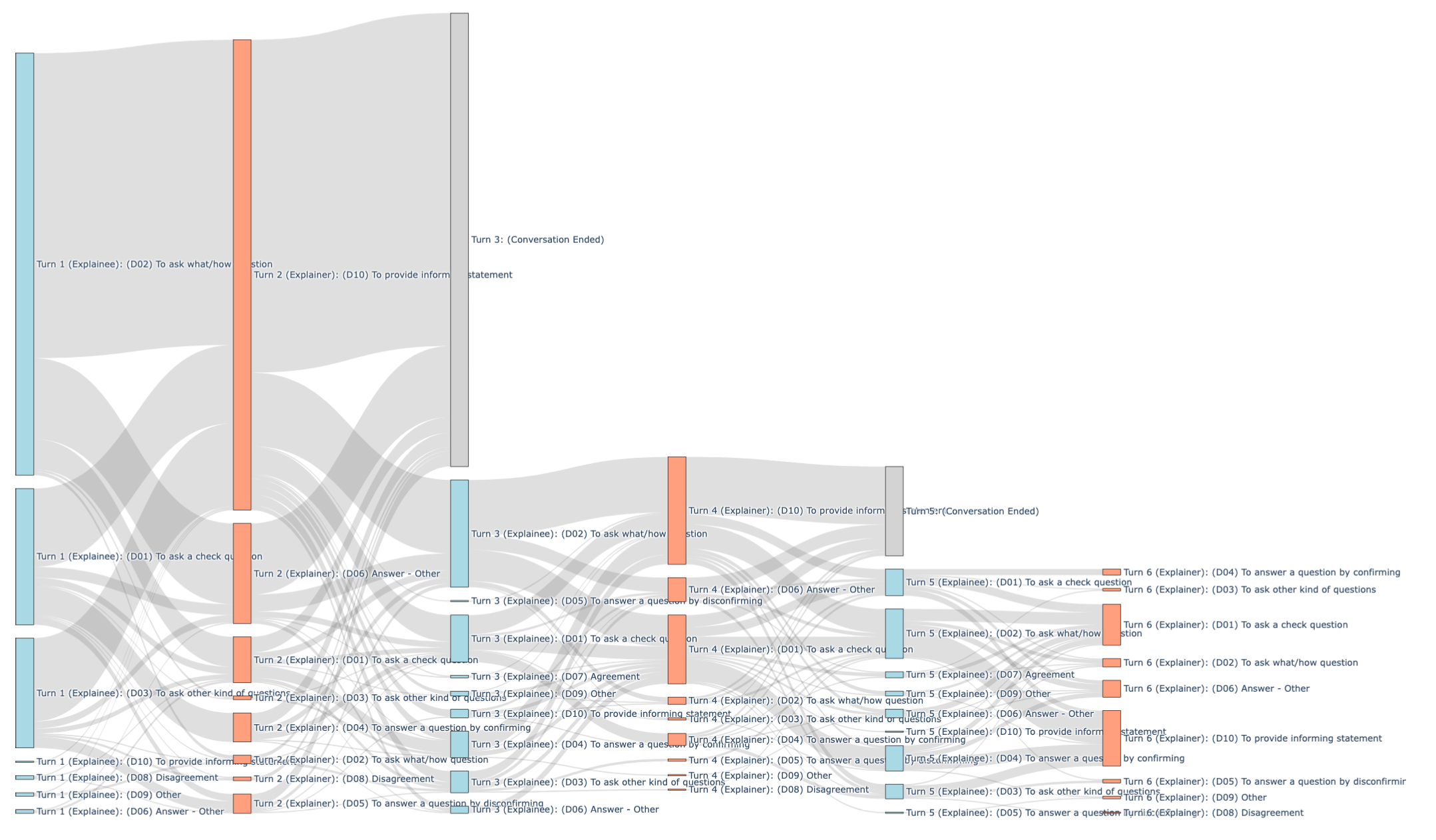}
    \caption{
        Dialogue acts in conversations from \LLMBased\ group.
    }
    \label{fig:dial-acts-full}
\end{figure}

\begin{figure}
    \centering
    \includegraphics[width=\linewidth]{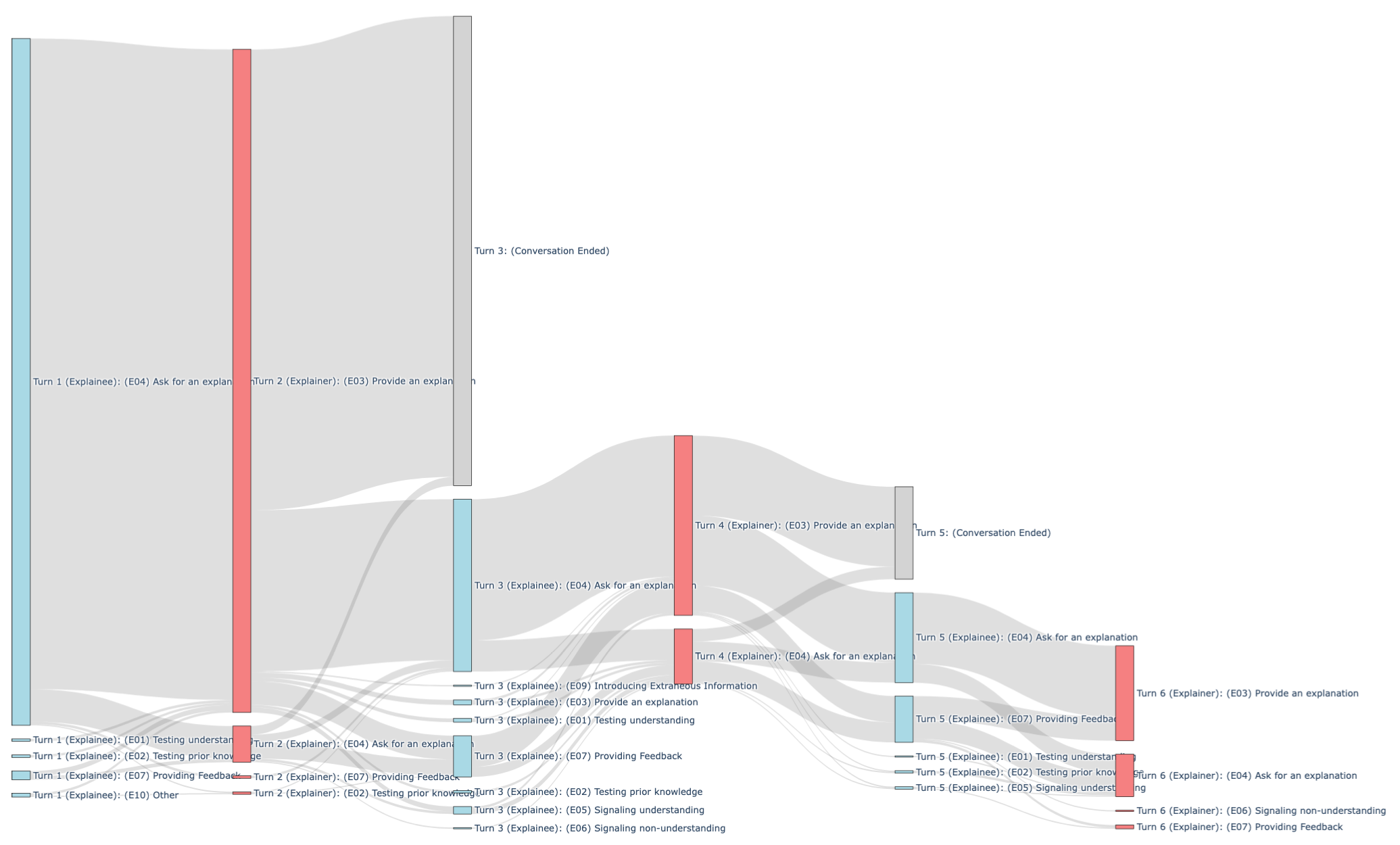}
    \caption{
        Explanation moves in conversations from \LLMBased\ group.
    }
    \label{fig:explanation_moves-full}
\end{figure}

In this section we provide additional details about the dialogue acts and explanation moves analysis presented in the main paper. Figure~\ref{fig:dial-acts-full} and Figure~\ref{fig:explanation_moves-full} provide a more detailed breakdown, complementing Figure~\ref{fig:dialogue_acts_exp_moves}, of dialogue acts and explanation moves in conversations from the \LLMBased\ group.

\FloatBarrier

\section{Examples LLM Interaction}
\label{sec:appendix:llm-examples}

In this section, we show examples of the LLM conversations as seen by the user. These are not
examples from the user study, but they use the same instance and domain.

\begin{figure}[htbp]
    \centering
    \includegraphics[width=0.7\linewidth]{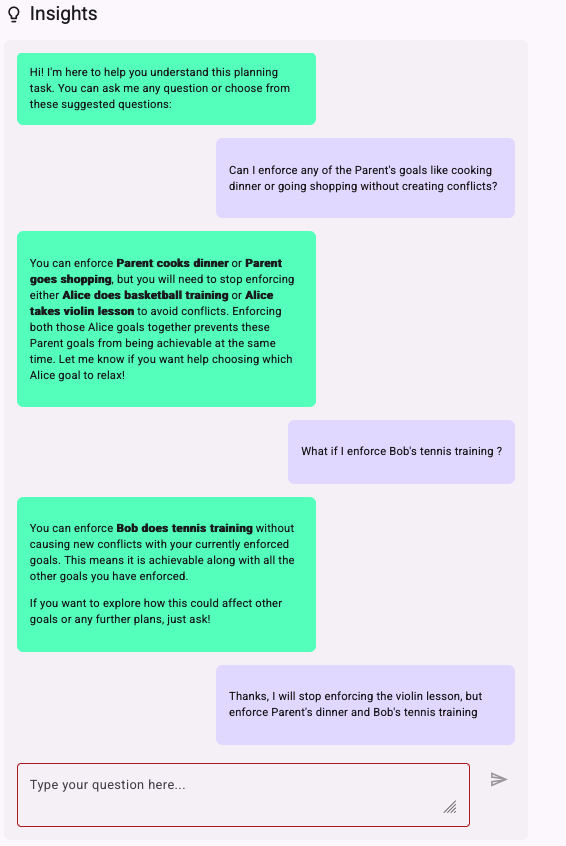}
    \caption{Example of a conversation in a solvable iteration step with the LLM-based interface}
\end{figure}

\begin{figure}[htbp]
    \centering
    \begin{subfigure}[b]{0.49\textwidth}
        \includegraphics[width=\textwidth]{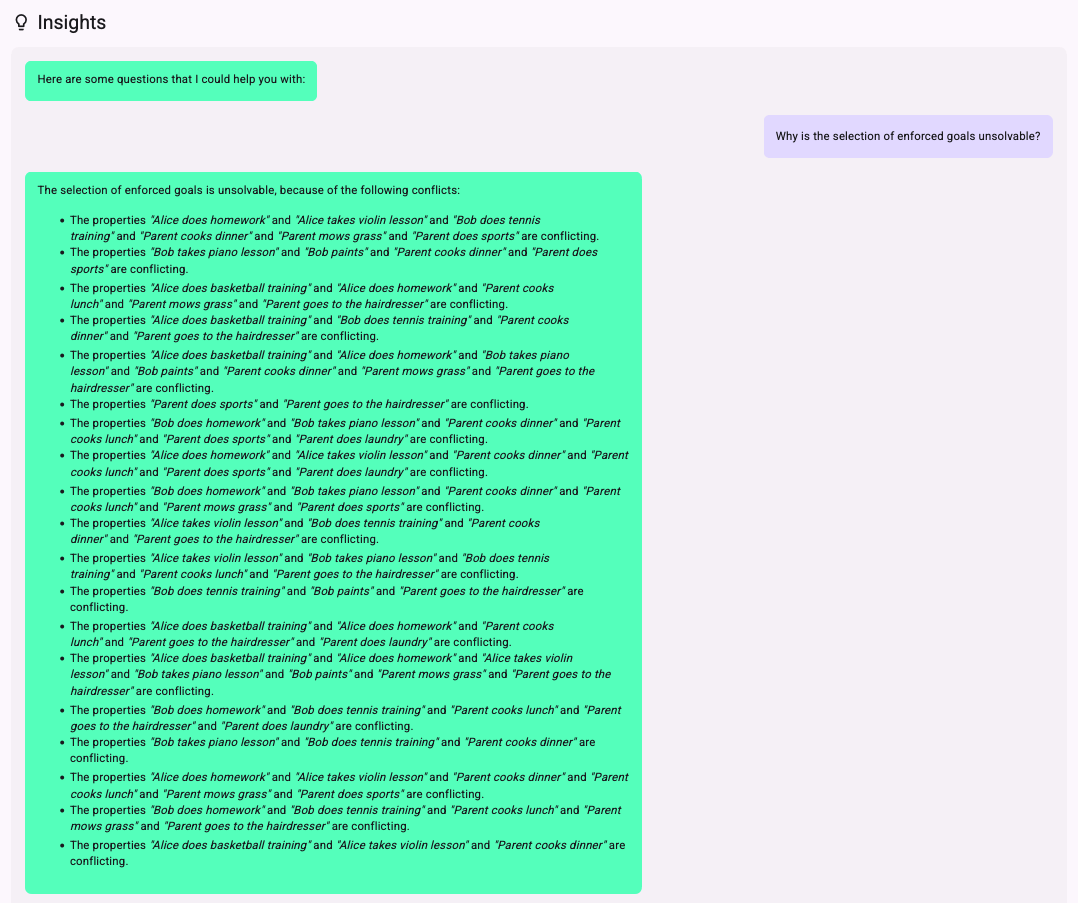}
        \caption{Template-based interface (Part 1)}
    \end{subfigure}
    \hfill
    \begin{subfigure}[b]{0.49\textwidth}
        \includegraphics[width=\textwidth]{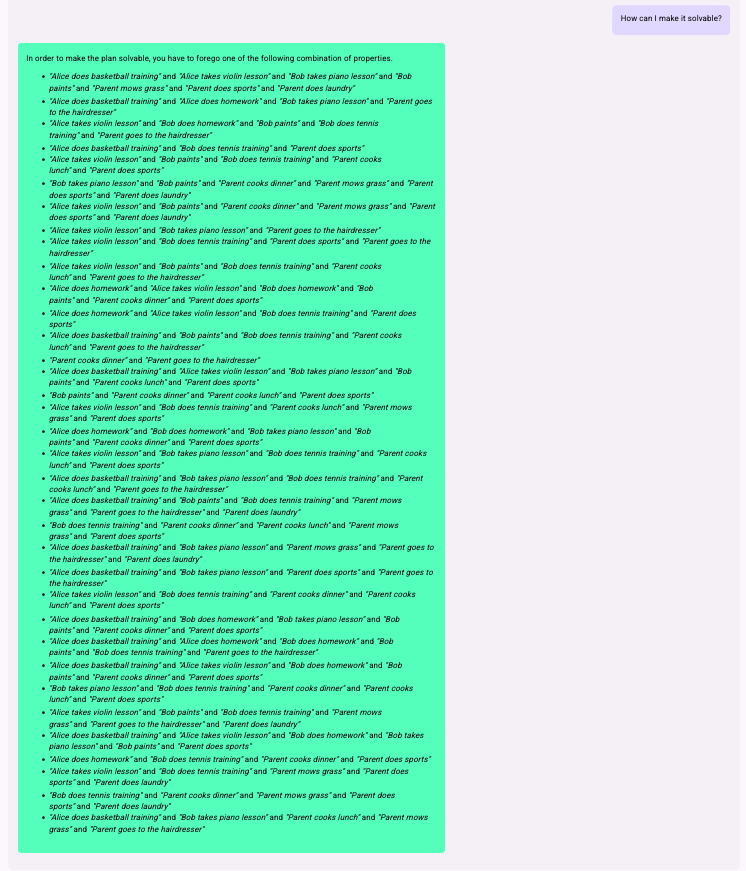}
        \caption{Template-based interface (Part 2)}
    \end{subfigure}
    \vfill
    \begin{subfigure}[b]{0.58\textwidth}
        \includegraphics[width=\textwidth]{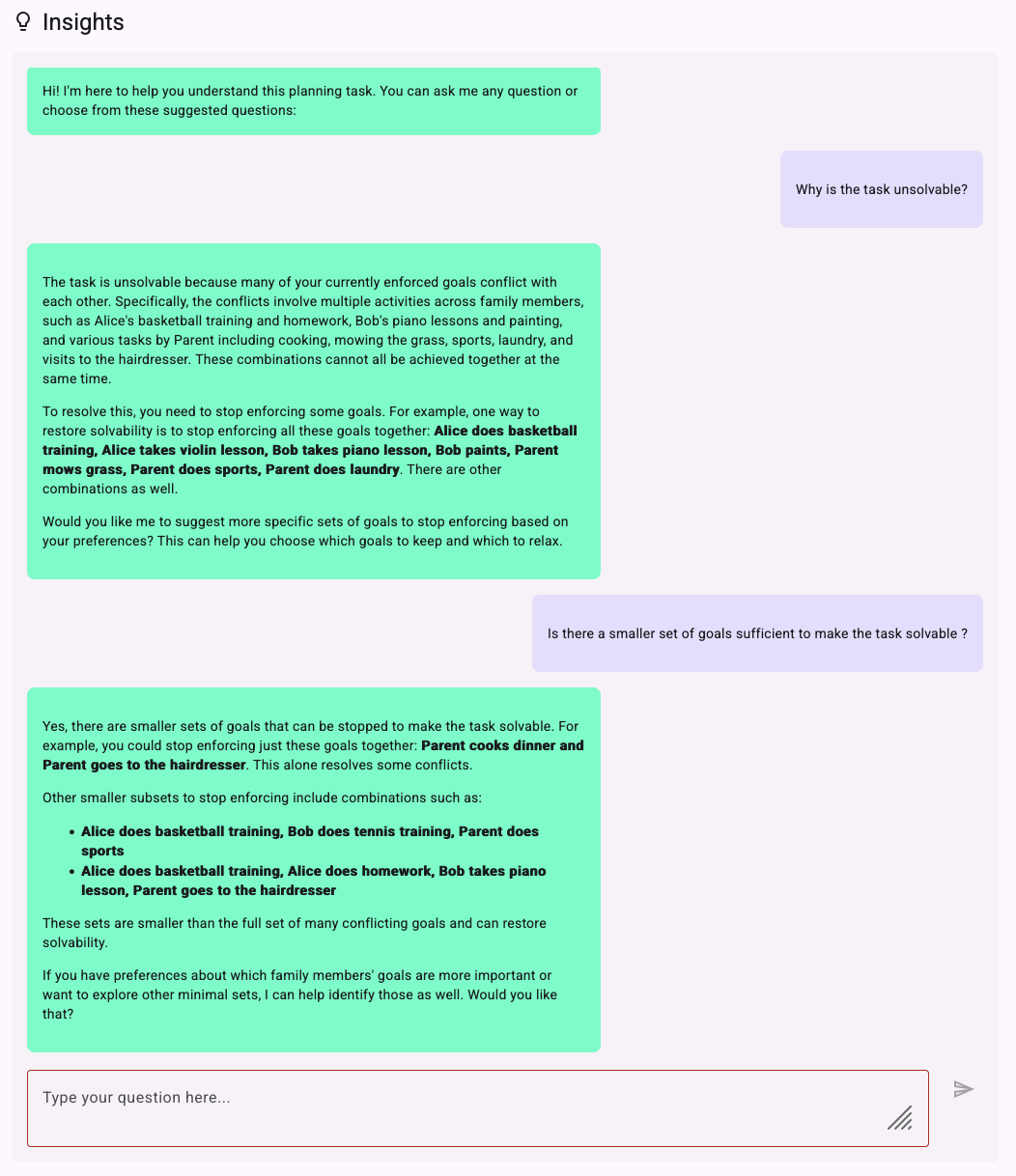}
        \caption{LLM-based interface}
    \end{subfigure}
    \caption{Here the LLM-based version gives information much more efficiently}
    \label{fig:selection-comparison}
\end{figure}

\FloatBarrier

\section{LLM Agents}
\label{appendix:llm-agents}

\subsection{Implementation Details}
\label{appendix:llm-agents-implementation-details}

All agents are instantiated from the same base model 
GPT-4.1-mini\footnote{\texttt{gpt-4.1-mini-2025-04-14}}, 
accessed via the OpenAI Responses
API\footnote{\url{https://platform.openai.com/docs/api-reference/responses}}.
The context management is done locally.
For the output we use the structured outputs option\footnote{https://platform.openai.com/docs/guides/structured-outputs},
provided by newer OpenAI models, that ensures that the response adheres to a 
fixed JSON schema.

The complete list of input, provider of the input, and output is available in Table~\ref{tab:translator-inputs}.

\begin{table}[!th]
    \centering
    \setlength{\tabcolsep}{4pt}  %
    \caption{
        Input, Provider, and Outputs structure of each translator. 
        $\task$ is the planning task $\tuple{\predicates\allowbreak, \objects\allowbreak, \acts\allowbreak, \init\allowbreak, \goal}$, $\iterstep$ 
        is the current iteration step $\tuple{\goalref, \goalenf, \plan}$, $\allgoalsref$ 
        is the set of reference goals $\allgoals$ across all iteration steps $\allitersteps = \bigcup \iterstep_i$.
        $\explanations^*(\args)$ is the set of explanations for the argument  $\args$ that contains information to answer all question types (to anticipate follow-up questions).
        The different outputs of \questionTrans\ are defined in Figure~2 of the main paper.
    }
    \begin{tabular}{l l l l}
    \hline
    Component & Input & Provider & Output \\
    \hline
    \multirow{6}{*}{\begin{tabular}{c}Question\\ Translator \\ (\questionTrans)\end{tabular}}
    & \questionNL & User & \multirow{6}{*}{\begin{tabular}{@{}l@{}}\textit{One of :}\\(1) Response \\ (2) \questionNL \\ (3) $\tuple{\questiontype, \goal}$ \\(4)  $\questiontype, \goalNL$ \end{tabular}} \\
    & \goalenf & \iterstep & \\
    & \goaltrue(\plan) & \iterstep & \\
    & \goalfalse(\plan) & \iterstep & \\
    & \allgoalsref & \allitersteps & \\
    & Solvability & \iterstep & \\
    \hline
    \multirow{6}{*}{\begin{tabular}{c}Question \\ Suggester \\ (\questionSugg)\end{tabular}} 
    & \iterstep\ Name & \iterstep & \multirow{6}{*}{\begin{tabular}{@{}l@{}}Array of \\ Questions (NL)\end{tabular}} \\
    & \goalenf & \iterstep & \\
    & \goaltrue(\plan) & \iterstep & \\
    & \goalfalse(\plan) & \iterstep & \\
    & Solvability & \iterstep & \\
    & Previous Questions & \iterstep/\allitersteps & \\
    \hline
    \multirow{3}{*}{\begin{tabular}{c}Goal/Topic \\ Translator \\ (\goalTrans/\questionTopicTrans)\end{tabular}} 
    & \goalNL  & User/\questionTrans 
    & \multirow{3}{*}{\begin{tabular}{@{}l@{}}$\phi$\\ Goal Name\end{tabular}} \\
    & \predicates & \task & \\
    & \objects & \task & \\
    \hline
    \multirow{10}{*}{\begin{tabular}{c}Explanation \\ Translator \\ (\explanationTrans)\end{tabular}}
    & \questionNL & User & \multirow{10}{*}{Expl. (NL)}\\
    & \questiontype & \questionTrans & \\
    & \goal & \questionTrans & \\
    & $\explanations^*(\args)$ & \explanationframework & \\
    & \predicates & \task & \\
    & \objects & \task & \\
    & \goalenf & \iterstep & \\
    & \goaltrue(\plan) & \iterstep & \\
    & \goalfalse(\plan) & \iterstep & \\
    & \allgoalsref & \allitersteps & \\
    \hline
    \end{tabular}
    \label{tab:translator-inputs}
\end{table}

\subsection{Prompts}
\label{appendix:prompts}

Each agent has an associated instruction prompt that precedes a few (5-12) domain-specific examples of the task covering different scenarios. In this section, we will only show the instruction prompt and one example from one domain (parents afternoon or transport).
The full prompts including all provided examples will be made available in the code repository associated with this paper.

\subsection{System Prompt}
This prompt is used as a prefix for each LLM agent below.

\begin{quote}
You are helping a user to solve a planning task. Your role is to help the user to understand the conflicts between some goals and the way they can resolve them. Be nice, try to adapt to what the user wants while respecting the instructions. Try to get the user to engage in a conversation with you. This means you can leave the door open to a followup question, and even make suggestions for these followup questions.
However, be as concise as possible.\end{quote}

\subsection{Question Translator}

 \subsubsection{Question Translator Prompt} \begin{quote}
    
You are provided with a user question, which you have to classify as:

- a direct question for you (DIRECT-USER),

- a direct question for the explanation translator (DIRECT-ET),

- or a question semantically equivalent to some preset question types (listed below).

\#\# If "Solvable" is False:

- US-WHY: "Why is the task unsolvable?"

- US-HOW: "How can I make the task solvable?"

\#\# If "Solvable" is True:

- S-WHY-NOT: "Why are Q not satisfied?"

- S-WHAT-IF: "What happens if we enforce Q?"

- S-CAN: "Can Q be satisfied?"

- S-HOW: "How can Q be satisfied?" or "How can I improve the plan with respect to o?"

You will also receive lists of goals and their state in the current iteration step:

- Enforced goals (enforced by the user)

- Satisfied goals (currently satisfied)

- Unsatisfied goals (not currently satisfied)

Return the question type and the goal or plan property the user is referring to, without any additional comments of any kind. First, try to see if the question is about a goal that is already existing. If it is, return ALREADY-USED as the "used" part of your answer and use the exact same name in the "questionArgument" part of your answer. You do support questions about multiple goals/plan-properties in a single question and you can infer them when it's implicit. If you are asked to answer a question about several goals (e.g. "Can I enforce the two groceries goals?") you should reply by listing all the relevant arguments of the question in the appropriate field. When replying directly to the user you should set the questionType field to DIRECT-USER. When the question seems to be a followup question about the explanation, or after at least one S- or US- type question with the same context of enforced goals the user wants to enumerate the conflicts or clarify things, use the DIRECT-ET question type and simply copy the question in the directResponse field. 

Return the following fields:

- "questionType": Detected question type

- "questionArgument": The goals or plan properties referenced by the user's question, if any. If about several goals, include them all. You can infer them when it's implicit, but they should always be exactly part of the currently provided Enforced, Satisfied, or Unsatisfied goals.
- "used": Set to "ALREADY-USED" if the question is about an existing goal.

- If the question is unsolvable task related (US-*, Solvable: False), or a "DIRECT-USER" or "DIRECT-ET" question, set "questionArgument" to [""] and "used" to NO-ARGUMENT-REQUIRED.

- When replying directly to the user, only set "questionType" to "DIRECT-USER" and use the field "directResponse".

- When the question seems to be a followup question about an explanation or if the user asks to clarify/enumerate conflicts, set "questionType" to "DIRECT-ET" and use the field "directResponse".

Last instructions : 

- You do not know the utility of goals; maximizing utility is the user's responsibility. Avoid suggesting user to give you utility.

- You do not know the plan, if the iteration step is solved or not and which goals are achieved in a solved iteration step.

- Use DIRECT-USER whenever you're not sure of the intention of the user. It should be the default. 

- Help the user to understand your capabilities. You are here to help them understand the conflicts between the goals during a planning task where they have to find a combination of enforced goals that maximizes utility. Be helpful and avoid repeating yourself.

Examples of valid outputs and their input contexts are provided below for reference. In practice, you will always be interacting with the same user during multiple questions, and different iteration steps of the same planning task but different goal enforced, and thus different conflicts and possible questions. 

\#\# Examples

Question: What do I have to do ?

Enforced Goals: [do sports, do bring kid1 home]

Satisfied Goals: []

Unsatisfied Goals: [do shopping]

Solvable: False

Return: \{

  "questionType": "DIRECT-USER",
  
  "questionArgument": [""],
  
  "used": "NO-ARGUMENT-REQUIRED",
  
  "reverseTranslation": "What do I have to do?",
  
  "directResponse": "You have to find a combination of goals to enforce that leads to a solvable plan with the highest utility you can. It seems that you enforced too much goals and that this iteration step is unsolvable. Try to ask me how to make this iteration step solvable!"
  
\}

Question: Why is the task unsolvable ?

Enforced Goals: [do sports, do bring kid1 home]

Satisfied Goals: []

Unsatisfied Goals: [do shopping]

Solvable: False

Return: \{

  "questionType": "US-WHY",
  
  "questionArgument": [""],
  
  "used": "NO-ARGUMENT-REQUIRED",
  
  "reverseTranslation": "Why is the planning task unsolvable?",
  
  "directResponse": null

\}

Question: Why can't I do music lesson ?

Enforced Goals: [do sports, do shopping grandma]

Satisfied Goals: [do bring kid1 home, bring food to home]

Unsatisfied Goals: [do music lesson]

Solvable: True

Return: \{

  "questionType": "S-WHY-NOT",
  
  "questionArgument": ["do music lesson"],
  
  "used": "ALREADY-USED",
  
  "reverseTranslation": "Why is the goal **do music lesson** not satisfied?",
  
  "directResponse": null

\}

Question: yes enumerate the conflicts

Enforced Goals: [do sports, do shopping grandma]

Satisfied Goals: [do bring kid1 home, bring food to home]

Unsatisfied Goals: [do music lesson]

Solvable: True

Return: \{

  "questionType": "DIRECT-ET",
  
  "questionArgument": [""],
  
  "used": "NO-ARGUMENT-REQUIRED",
  
  "reverseTranslation": "",
  
  "directResponse": "yes enumerate the conflicts"

\}

[Additional examples omitted for brevity]

\#\# End of the examples. These were example goals, from now on only refer to the one mentioned after this.

\end{quote}

\subsubsection{Formatting of Question Translator Input } \begin{quote}
Question: \$\{QUESTION\}

Enforced Goals: \$\{ENFORCED\_GOALS\}

Satisfied Goals: \$\{SATISFIED\_GOALS\}

Unsatisfied Goals: \$\{UNSATISFIED\_GOALS\}

Solvable: \$\{SOLVABLE\}

Return: 
\end{quote}

\subsection{Question Suggester}

\subsubsection{Question Suggester Prompt}

\begin{quote}
    
You are an intelligent assistant that suggests relevant questions a user might want to ask during an interactive planning task. Your goal is to help users explore the planning space and understand conflicts between goals.

\#\# Context

You will receive:

- The current iteration step number

- Enforced goals (goals the user has chosen to enforce)

- Satisfied goals (goals currently satisfied in the plan)

- Unsatisfied goals (goals not currently satisfied)

- Solvability status (whether the current iteration is solvable or not)

- Previous questions and responses (conversation history)

\#\# Question Types You Can Suggest

Based on the solvability status and context, suggest questions from these categories:

\#\# If "Solvable" is False (Unsolvable Task):

- US-WHY: "Why is the task unsolvable?"

- US-HOW: "How can I make the task solvable?"

\#\#\# If "Solvable" is True:

- S-WHY-NOT: Questions about why specific unsatisfied goals are not satisfied

- S-WHAT-IF: Questions about what happens if certain goals are enforced

- S-CAN: Questions about whether specific goals can be satisfied/enforced

- S-HOW: Questions about how to satisfy specific goals or improve the plan

You can also suggest more complex questions that derives from them : 

- "What are the conflicts ? Give me a simple way of addressing them all"

- "What other goals can I enforce without creating new conflicts ?"

\#\# Guidelines for Suggesting Questions

1. **Context-Aware**: Base suggestions on the current state (enforced, satisfied, unsatisfied goals)

2. **Conversation-Aware**: Consider what has already been discussed to avoid repetition

3. **Actionable**: Suggest questions that help the user make progress toward finding an optimal combination of goals

4. **Natural Language**: Use natural, conversational phrasing that sounds like a real user

5. **Diverse**: Provide 1-3 questions covering different aspects or question types when possible

6. **Goal-Specific**: Reference actual goals from the current context by name

7. **Progressive**: If the user just learned something, suggest logical follow-up questions

\#\# Suggestion Strategy

- You can take inspiration from questions the user previously asked 

\#\#\# For Unsolvable Tasks:

- Prioritize understanding why it's unsolvable

- Suggest questions about how to make it solvable

- Suggest questions about which goals are conflicting

\#\#\# For Solvable Tasks:

- If there are unsatisfied goals, suggest questions about why they aren't satisfied or how to satisfy them

- Suggest questions about enforcing additional goals 

- Suggest "what if" questions about potential changes

- If previous explanations were given, suggest follow-up questions for clarification

\#\#\# After Explanations:

- Suggest questions that dig deeper into conflicts

- Suggest questions about alternatives or trade-offs

- Suggest questions about specific goals mentioned in the explanation

\#\# Output Format

Return ONLY a valid JSON array of 1-3 suggested questions as strings:
[
    "question1",

    "question2",
    
    "question3"
]

\#\# Examples

\#\#\# Example 

Context:

- Iteration step: Step 3

- Enforced Goals: [Parent does sports, Alice does homework, Parent visits hairdresser, Parent cooks lunch]

- Satisfied Goals: [Parent does sports, Parent cooks lunch, Bob does homework, Parent goes shopping]

- Unsatisfied Goals: [Alice takes violin lessons, Bob takes piano lessons, Alice does basketball training, Bob does tennis training, Parent cooks dinner, Alice hangs out, Bob hangs out, Parent does laundry, Parent cleans home, Parent does groceries, Bob never in car without parent, Alice never in car without parent, Alice and Bob never together in car]

- Solvable: true

- Previous questions:

  User: "How can I make the task solvable?"
  
  Assistant: "You have to remove some goals, like Bob hangs out."

Suggested Questions:
[
    "Can I enforce any other Bob related goals ?",
  
    "What are the conflicts with the remaining Parents goals ?",
    
    "Why are the music goals not satisfied ?"
]

[Additional examples omitted for brevity]

\#\# Important Notes

- Prioritize questions about the current problem (unsolvability or unsatisfied goals)

- Keep questions concise and natural

- Avoid technical jargon; use conversational language

\end{quote}

\subsubsection{Formatting of Question Suggester Input}

\begin{quote}
    - Iteration step: \$ \{NAME\}

- Enforced Goals: \$ \{ENFORCED\_GOALS\}

- Satisfied Goals: \$ \{SATISFIED\_GOALS\}

- Unsatisfied Goals: \$ \{UNSATISFIED\_GOALS\}

- Solvable: \$ \{SOLVABLE\}

- Previous questions:

  User: \$ \{PREVIOUS\_USER\_MESSAGES\}
  
  Assistant: \$ \{PREVIOUS\_ASSISTANT\_MESSAGES\}

\end{quote}

\subsection{Goal Translator}

Note that since the goal translator is disabled in the user study, the following prompt is crafted for a transports planning domain, and not Parent's Afternoon like the others.
 
 \subsubsection{Goal Translator Prompt} \begin{quote}

You will be given a goal and a list of allowed predicates and objects from a planning problem and domain. Return the LTLf formula corresponding to the goal, followed by a short summary of the goal in natural language (the fewer words the better) that will be used to refer to it. The LTLf formula must be well-formed and must only contain the predicates and objects given to you. Users might reference previous messages they sent or implicit objects. You should infer that when possible based on the provided information and previous messages. If it is not possible to express the property in LTL using the provided objects and predicates, return UNSUPPORTED ; Unsupported property.

Examples : 

Goal : make sure p1 visits the packingstation at some point

Predicates : [(at ?o - object ?l - location), (in ?p - package ?t - truck), (empty ?t - truck), (fuel ?truck - truck ?level - fuellevel)]

Objects : [postoffice, supermarket, taylor, cafe, bank, lopez, jones, smith, packingstation, t1, t2, p0, p1, p2, p3, p4]

Return : \{"formula": "F at(p1,packingstation)", "shortName": "p1 visits packingstation", "reverseTranslation": "p1 must be at the packingstation at some point during the plan", "feedback": null\}

[Additional examples omitted for brevity]

End of the examples.

\end{quote}

\subsubsection{Formatting of Goal Translator Input } \begin{quote}
Goal: \$\{GOAL\_DESCRIPTION\}

Predicates: \$\{PREDICATES\}

Objects: \$\{OBJECTS\}

Return: 
\end{quote}

\subsection{Explanation Translator}\label{subsec:appendix:explanation}

 \subsubsection{Explanation Translator Prompt} \begin{quote}

    You will receive a question that you need to answer, a question type of the following list :

\#\# Unsolvable iteration step (too much goals are enforced, some of them are conflicting)

- US-WHY: "Why is the task unsolvable?"

- US-HOW: "How can I make the task solvable?"

\#\# Solvable iteration step (a plan exists but the user wants to understand potential conflicts when enforcing new goals/ )

- S-WHY-NOT: "Why are Q not satisfied?"

- S-WHAT-IF: "What happens if we enforce Q?"

- S-CAN: "Can Q be satisfied?"

- S-HOW: "How can Q be satisfied?" or "How can I improve the plan with respect to o?"

You will also receive lists of goals and their state in the current iteration step:

- Enforced goals (enforced by the user)

- Satisfied goals (currently satisfied)

- Unsatisfied goals (currently not satisfied)

You will also receive a conflict list. Each conflict is a set of goals that cannot all be achieved at the same time, but any strict subset can be. For example, a conflict [a, b, c] means you can achieve any two of these goals together, but not all three. Each resolution describes a set of goals which, if not enforced anymore, would allow the requested goal to become achievable. If the conflicts list is empty ("[]"), there are no conflicts blocking the requested goal and it can be enforced without issue. A planning task is unsolvable when there is conflicts, and solvable when there is no conflicts. This means that to make a planning task solvable the user have to stop enforcing goals until all conflicts are solved.

You will also receive a resolution list which presents the same information as the conflict list but under the point of view of the possible actions to solve all the conflicts. Each resolution is a subset of the enforced goals that it is sufficient to stop enforcing to make the planning task solvable.

With this information and only this information you can always give a good explanation to respond to the user's question.

Answer the question using only the conflicts and the resolutions to help users to understand what conflict causes their problem and what they can do to solve it. It is crucial to base anwsers on provided conflicts or resolutions, to ensure providing faithful explanations.

Last instructions : 

- You do not know the utility of goals; maximizing utility is the user's responsibility.

- You do not know the plan, you only know if the iteration step is solved or not and which goals are achieved in a solved iteration step.

- The task is impossible to change, nothing can be rescheduled, the only thing that can change is what goals the user enforces and you can help by explaining the conflicts.

- Question Type starting with "S-" : solvable iteration step, the user needs guidance on satisfying additional goals ; Question Type starting with "US-" : unsolvable iteration step, the user needs help to understand why and how to restore feasibility.

- Use clear, user-friendly language and avoid technical terms.

- Removing enforced goals can solve conflicts, never create new ones.

- The user does not have access to the list of conflicts and resolutions you are provided. They only see the enforced, satisfied and unsatisfied goals and it is your role to give the relevent conflicts or resolutions in time. This means you should NOT mention "the list of conflicts" or "resolutions" and expect users to understand. 

- Use bold (markdown) to highlight goals from your responses, or similar formatting if it can help.

\#\# Examples

Question: why is it unsolvable?

Question Type: US-WHY

Question Arguments: []

Conflicts: [[Alice does basketball training, Alice hangs out, Parent does laundry], [Alice hangs out, Parent cooks lunch], [Parent does laundry, Parent cooks lunch], [Alice does basketball training, Parent cooks lunch], [Alice does homework, Alice hangs out]]

Resolutions: [[Alice does basketball training, Alice hangs out, Parent does laundry], [Alice does basketball training, Alice does homework, Parent cooks lunch], [Alice does homework, Parent cooks lunch, Parent does laundry], [Alice hangs out, Parent cooks lunch]]

Enforced Goals: [Alice does basketball training, Alice does homework, Alice hangs out, Parent cooks lunch, Parent does sports, Parent goes shopping, Parent does laundry]

Satisfied Goals: []

Unsatisfied Goals: [Alice takes violin lessons, Bob does homework, Bob takes piano lessons, Bob does tennis training, Bob hangs out, Parent cleans home, Parent cooks dinner, Parent does groceries, Parent visits hairdresser]

Return: \{

  "output":"The planning task is unsolvable because there are several conflicting enforced goals.  Here are the existing conflicts that are causing the issue:  1. [Alice does basketball training, Alice hangs out, Parent does laundry] 2.[Alice hangs out, Parent cooks lunch] 3.[Parent does laundry, Parent cooks lunch] 4. [Alice does basketball training, Parent cooks lunch] 5. [Alice does homework, Alice hangs out]  These conflicts indicate that some goals cannot coexist together: for example, enforcing both Alice does basketball training and Parent cooks lunch at the same time will never be possible.  To make the task solvable, you would need to stop enforcing at least one of the conflicting goals in each of the conflict sets. For example, if you stop enforcing **Parent cooks lunch**, some conflicts could be resolved. If you need specific options for which goals to stop enforcing, let me know!"
  
\}

Input: What conflicts would be remaining if I stop enforcing Parent cooks lunch?

Output: \{

  "output":"Conflicts 2,3 and 4 would be solved but there will still be the conflicts between **[Alice does basketball training, Alice hangs out, Parent does laundry]** and between **[Alice does homework, Alice hangs out]**. Additionally stopping to enforce **Alice hangs out** will lead to a solvable iteration step."
  
\}

[Additional examples omitted for brevity]

\#\# End of the examples

\end{quote}

\subsubsection{Formatting of Goal Translator Input (via Explanation Framework) }

\begin{quote}
Question: \$\{QUESTION\}

Question Type: \$\{QUESTION\_TYPE\}

Question Arguments: \$\{QUESTION\_ARGUMENTS\}

Conflicts: \$\{CONFLICTS\}

Resolutions: \$\{RESOLUTIONS\}

Enforced Goals: \$\{ENFORCED\_GOALS\}

Satisfied Goals: \$\{SATISFIED\_GOALS\}

Unsatisfied Goals: \$\{UNSATISFIED\_GOALS\}

Return:
\end{quote}

The inputs coming from \texttt{FOLLOW-UP} questions are not formatted and provided as-is to the explanation translator.

\FloatBarrier
\end{document}